\documentclass{article} 
\usepackage{iclr2026_conference,times}

\usepackage{amsmath,amsfonts,bm}

\def\eqref#1{equation~\ref{#1}}

\def\1{\bm{1}}

\DeclareMathAlphabet{\mathsfit}{\encodingdefault}{\sfdefault}{m}{sl}
\SetMathAlphabet{\mathsfit}{bold}{\encodingdefault}{\sfdefault}{bx}{n}

\usepackage{url}

\iclrfinalcopy 

\usepackage{multibib}

\newcites{sup}{Supplementary References}

\usepackage[utf8]{inputenc} 
\usepackage[T1]{fontenc}    
\usepackage{url}            
\usepackage{booktabs}       
\usepackage{amsfonts}       
\usepackage{nicefrac}       
\usepackage{microtype}      
\usepackage[dvipsnames]{xcolor}         

\usepackage{amsmath}
\usepackage{wrapfig}
\usepackage{caption}
\usepackage{multirow} 
\usepackage{adjustbox}
\usepackage{enumitem}

\newcommand{\multiset}[1]{
  \{\!\!\{#1\}\!\!\}
}
\newcommand{\difflift}{\partial \mathrm{lift}}
\usepackage{bbm}
\usepackage[colorlinks=True, linkcolor=violet, citecolor=violet]{hyperref}

\newtheorem{remark}{Remark}

\definecolor{lb}{RGB}{31,119,180}

\usepackage{tcolorbox}
\newtcolorbox{mybox}[1]{colback=lb!1!white,colframe=lb!70!black,fonttitle=\bfseries,title=#1}

\title{Differentiable Lifting for Topological Neural Networks}

\author{%
  Jorge Luiz Franco \\
  University of São Paulo, Instituto Curvelo \\
  \And
  \hspace{-0pt}Gabriel Duarte \\
  \hspace{-0pt}Federal Institute of Ceará \\
  \And
  Alexander Nikitin \\
  Aalto University \\
  \And
  Moacir Ponti \\
  University of São Paulo \\
  \And
  Diego Mesquita \\
  Getulio Vargas Foundation, $2\delta$ AI \\
  \And
  Amauri H. Souza \\
  Federal Institute of Ceará, $2\delta$ AI \\
}

\begin{document}

\maketitle

\begin{abstract}
Topological neural networks (TNNs) enable leveraging high-order structures on graphs (e.g., cycles and cliques) to boost the expressive power of message-passing neural networks. In turn, however, these structures are typically identified \emph{a priori} through an unsupervised graph lifting operation. Notwithstanding, this choice is crucial and may have a drastic impact on a TNN's performance on downstream tasks. To circumvent this issue, we propose $\difflift$ (DiffLift), a general framework for learning graph liftings to hypergraphs and cellular- and simplicial complexes in an end-to-end fashion. In particular, our approach leverages learned vertex-level latent representations to identify and parameterize distributions over candidate higher-order cells for inclusion. This results in a scalable model which can be readily integrated into any TNN. Our experiments show that $\difflift$ outperforms existing lifting methods on multiple benchmarks for graph and node classification across different TNN architectures. Notably, our approach leads to gains of up to 45\% over static liftings, including both connectivity- and feature-based ones.

\end{abstract}

\section{Introduction}
\vspace{-2pt}
Topological neural networks (TNNs) \citep{Papillon23,Bodnar2021,Verma2024} have recently emerged as a prominent class of models for learning on topological domains, such as hypergraphs and simplicial complexes, with many researchers arguing they represent the new frontier for relational learning~\citep{Papamarkou2024}. 
Akin to graph neural networks (GNNs)~\citep{scarselli2009,Gilmer2017}, typical TNNs employ message-passing layers where each element of the input (e.g., nodes or cells) updates its representation (features) based on those of its topological neighbors. 
Thus, these models generalize convolution-like operations on graphs to higher-order relational objects.
Importantly, the primary application of TNNs has been to enhance the capabilities of graph-based models, particularly in terms of expressivity~\citep{Bodnar2021,bodnar2021weisfeiler}. In this context, the input graphs must first be transformed to the domain on which a TNN operates --- a process known as \emph{lifting}.
\looseness=-1

\captionsetup[figure]{font=small}
\begin{wrapfigure}[11]{r}{0.59\textwidth}
\vspace{-18pt}
  \centering
\includegraphics[height=0.20\textwidth]{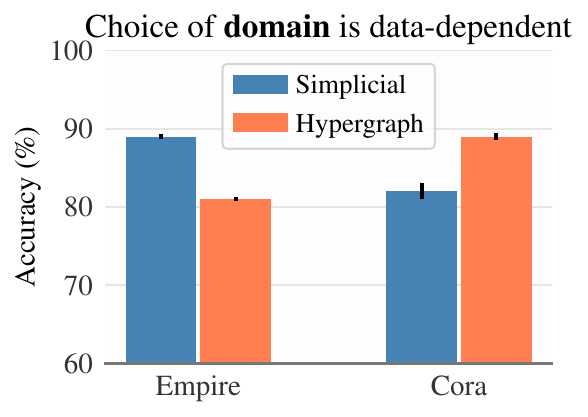} \includegraphics[trim=20pt 0pt 0pt 0pt, ,clip, height=0.21\textwidth]{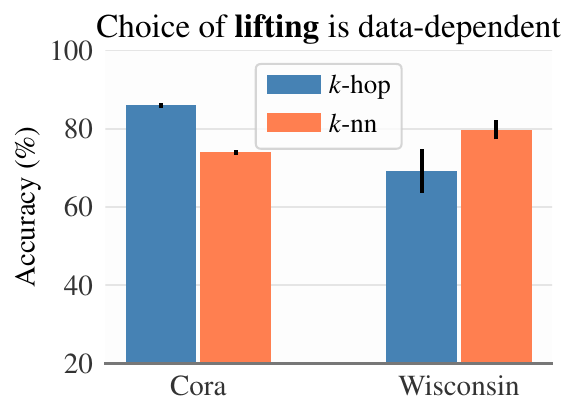}
\vspace{-6pt}
  \caption{[\textit{Left}] Lifting to different domains can lead to disparate performances. Accuracies taken from the best TNNs in \citep{telyatnikov2024topobenchmarkx}. [\textit{Right}] Performances of liftings to the same domain (hypergraph) vary greatly. Values taken from \autoref{tab:hypergraph_node}.}
    \label{fig:impact_lifting}
\end{wrapfigure}
\captionsetup[figure]{font=normal}

%


Lifting methods explore graph connectivity and features to create higher-order relational structures. 
For instance, \emph{clique lifting} \citep{bodnar2021weisfeiler} produces a simplicial complex by leveraging cliques in the input graph while
\emph{cycle lifting} \citep{hajij2022topological} detects cycles to create a cell complex.
In general, there are many lifting procedures for each topological domain --- c.f. Tab. 3 in \citet{telyatnikov2024topobenchmarkx}.

Not surprisingly, the optimal choice of topological domain and lifting procedure for each task is non-obvious, and its impact on performance is highly data-dependent. \autoref{fig:impact_lifting} compares TNNs on different domains, showing opposite behaviors depending on data, even within the same topological domain.  
\looseness=-1






Strikingly, despite the high impact of the lifting operation on TNNs, most lifting methods are not supervised and thus not informed by the task at hand \citep{hajij2022topological,telyatnikov2024topobenchmarkx}, which may lead to suboptimal architectures. To date, differentiable lifting has only been explored in the context of cell complexes \citep{battiloro2023latent}.
\looseness=-1

\begin{figure}[t]
    \centering
    \includegraphics[width=\linewidth]{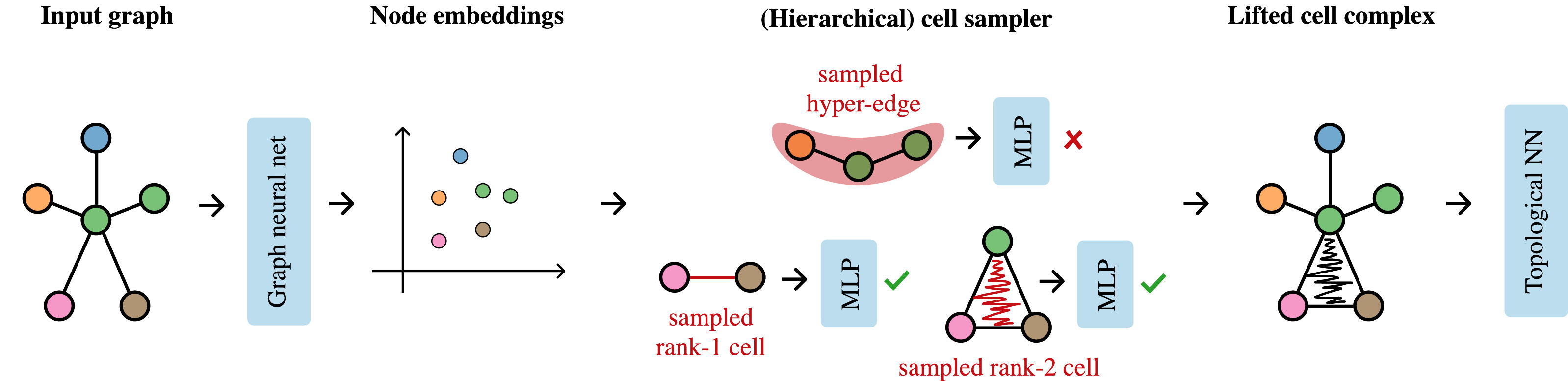}
    \caption{Overview of $\difflift$. For a given input graph, we first compute node embeddings using GNNs. Then, we use these embeddings to select cells/hyperedges. Cell-level embeddings run through MLPs responsible for returning acceptance probabilities. For hierarchical domains (e.g., cell complexes), cells are generated in increasing dimensionality. From the accepted cells, we form a relational object that is sent to an off-the-shelf TNN for graph/node-level predictions. The model is trained end-to-end.\looseness=-1}
    \label{fig:enter-label}
\end{figure}

This work proposes $\difflift$ (DiffLift) -- a general, differentiable lifting framework applicable to various domains, including hypergraphs and simplicial/cell complexes. Our method uses a probabilistic approach to sample candidate cells of adaptive sizes. Specifically, we parameterize distributions over cells using node embeddings derived from arbitrary graph models (e.g., GNNs or Graph Transformers \citep{gps_cite}). For each candidate cell, we compute its embedding and use a multilayer perceptron (MLP) to estimate the probability of accepting or rejecting the cell — that is, determining whether it should be included in the output structure. \autoref{fig:enter-label} 
provides a schematic overview of $\difflift$.
\looseness=-1

To model the typical hierarchical structure of topological objects, we propose an iterative sampling procedure, where cells are generated in increasing order of dimensionality: samples of dimension $i$ are used to inform the sampling of dimension $(i+1)$-cells. Notably, our approach generalizes across multiple topological domains and can be seamlessly integrated into standard TNN pipelines.
\looseness=-1

We evaluate $\difflift$ on 12 datasets spanning graph and node classification tasks using four different TNN models. Our results show that $\difflift$ consistently outperforms unsupervised lifting methods in nearly all graph-level classification benchmarks — achieving superior performance in 22 out of 24 experiments, often by a substantial margin. These gains are robust across all TNN architectures. For example, when using CW Networks \citep{Bodnar2021}, $\difflift$ yields performance gains of up to 45\%.
For node classification, $\difflift$ achieves competitive performance relative to static lifting methods and outperforms DCM \citep{battiloro2023latent} (a differentiable lifting baseline) overall. Additionally, we analyze the sensitivity of $\difflift$ to the choice of its GNN component, highlighting that while this choice often impacts the overall performance, our design is robust and produces strong empirical results even when adopting simple GNNs (e.g., graph isomorphism networks~\citep{xu2018how}).
 
\section{Background}

This section overviews the main types of relational structures and respective neighborhood notions, message-passing networks for relational data, and graph lifting methods. In the following, we assume readers are familiar with basic notions in topology; see \citep{munkres2000topology} for reference. 
\looseness=-1

\noindent \textbf{Graphs and hypergraphs.} We denote an \emph{undirected graph} as a tuple $G=(V, E)$ where $V$ is a set of vertices (or nodes) and $E$ is a set of unordered vertex pairs, i.e., edges. The set of neighbors of a node $v$ in $G$ is denoted by $\mathcal{N}^G(v)=\{u  \in V: \{v, u\} \in E\}$.
Hypergraphs generalize graphs by allowing edges to connect multiple nodes. Formally, a \emph{hypergraph} on a nonempty set $V$ is a pair $ (V, K)$, where $K \subseteq 2^V\setminus\emptyset$ and its elements are called hyperedges. 


\noindent \textbf{Simplicial complexes} are topological spaces comprised of simple mathematical objects called simplices (points (0-simplices), line segments (1-simplices), triangles (2-simplices), and their higher-dimensional analogues). In particular, an \emph{abstract simplicial complex} (ASC) over a vertex set $V$ is a set $K$ of subsets of $V$ (the \emph{simplices}) such that, for every $\sigma \in K$ and every non-empty $\tau \subset \sigma$, we have that $\tau \in K$.
Thus, we can define ASCs as a family of subsets $K \subseteq 2^V$ of $V$ that is closed under taking subsets. 
The dimension of a simplex is equal to its cardinality minus $1$, and the dimension of an ASC is the maximal dimension of its simplices.
We say $\tau$ is on the boundary of a simplex $\sigma$, denoted by $\tau \prec \sigma$, iff $\tau \subset \sigma$ and there is no $\delta$ such that $\tau \subset \delta \subset \sigma$, i.e., $\prec$ defines the boundary relation of $K$.
%
We note that undirected graphs correspond to 1-dimensional ASCs.
\looseness=-1

\begin{figure}[tb] 
  \centering
\includegraphics[width=\textwidth]{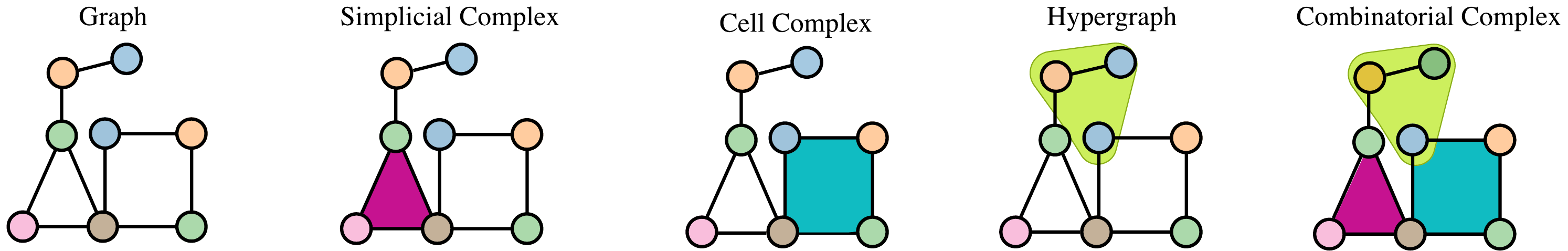}
  \caption{Examples of topological domains.}
    \label{fig:domains}

\end{figure}
\setlength{\textfloatsep}{5pt}
\noindent \textbf{Cell complexes.} A \emph{regular cell complex} \citep{Hansen2019} is a topological space $X$ with a partition $\{X_\sigma\}_{\sigma \in P_X}$ of subspaces $X_\sigma$ of $X$ called \emph{cells} such that
\begin{enumerate}[left=8pt,noitemsep,topsep=0pt]
    \item For each $x \in X$, there is an open neighborhood of $x$ that intersects finitely many cells;
    \item For all $\sigma, \tau \in P_X$, $X_\tau \cap \overline{X_\sigma} \neq \emptyset$ only if $X_\tau \subseteq \overline{X_\sigma}$, where $\overline{X_\sigma}$ denotes the closure of $X_\sigma$ (the intersection of all closed sets containing $X_\sigma$);
    \item Every cell $X_\sigma$ is homeomorphic to $\mathbb{R}^{D_\sigma}$ for some $D_\sigma$ which  we call $X_\sigma$'s dimension;
    \item For all $\sigma \in P_X$, there is a homeomorphism $\phi$ of a closed ball in $\mathbb{R}^{D_\sigma}$ to $\overline{{X}_\sigma}$ such that the restriction of $\phi$ to the interior of the ball is a homeomorphism onto $X_\sigma$.
\end{enumerate}

Importantly, the conditions (2) and (4) impose a poset structure $\tau \leq \sigma \iff X_\tau \subseteq \overline{X_\sigma}$ which fully characterizes the topology of the underlying cell complex $X$. This topological information can be described by the \emph{boundary relation} $\prec$ between two cells: $\sigma \prec \tau$ iff $\sigma < \tau$ and there is no cell $\delta$ such that $\sigma < \delta < \tau$, where $<$ denotes the strict version of the partial order $\leq$ above. 
We note that the class of cell complexes subsumes simplicial complexes. For more details on cell complexes, we refer to \cite{Hatcher02,Bodnar2021}. 
\looseness=-1

\noindent \textbf{Combinatorial complexes.} A \emph{combinatorial complex} (CC) \citep{hajij2022topological} is a tuple $(V, K, \mathrm{rk})$ where $V$ is a finite set, $K \subseteq 2^V\setminus \emptyset$ comprises a set of cells, and $\mathrm{rk}: K \rightarrow \mathbb{Z}_{\geq 0}$ is a ranking function s.t.\looseness=-1
\begin{enumerate}[left=8pt,noitemsep,topsep=0pt]
    \item For all $v \in V, \{v\} \in K$;
    \item For all $\sigma, \sigma' \in K, \sigma \subseteq \sigma' \implies \mathrm{rk}(\sigma) \leq \mathrm{rk}(\sigma')$.
\end{enumerate}
The idea of CCs is to generalize hierarchical structures (e.g., simplicial complexes) by  imposing mild relationships between cells via ranking functions --- CCs only require the order-preserving property in condition (2) --- while being flexible to accommodate non-hierarchical structures such as hypergraphs. \autoref{fig:domains} depicts the most popular relational structures in topological deep learning.
\looseness=-1

\noindent \textbf{Neighborhood structures.} We can exploit boundary relations (or rank functions) to specify local neighbors for each cell. In particular, \citet{bodnar2021weisfeiler} introduce four neighborhood structures:
\looseness=-1
\begin{itemize}[left=8pt,noitemsep,topsep=0pt]
    \item Boundary and co-boundary: $\mathcal{N}_B(\sigma)=\{\tau: \tau \prec \sigma \}$ and $\mathcal{N}_C(\sigma)=\{\tau:  \sigma \prec \tau \}$, respectively 
    \item Upper/lower adjacency:$\mathcal{N}_\uparrow(\sigma)\!=\!\{\tau : \exists \delta \text{ st }  \tau \prec \delta ,  \sigma \prec \delta\}$
    and $\mathcal{N}_\downarrow(\sigma)\!=\!\{\tau : \exists \delta \text{ st }  \delta \prec \tau , \delta \prec \sigma\}$
    \looseness=-1
\end{itemize}
Analogs of these neighborhoods can also be obtained via ranking functions \citep{hajij2022topological}.

\noindent \textbf{Features / signals.} In this work, we consider relational structures equipped with features. Let $K$ be a set of cells or hyperedges of a relational domain. Its attributed counterpart is a tuple ($K$, $x$) where $x: K \rightarrow \mathbb{R}^{d}$ assigns a feature vector $x(\sigma)$ to each cell $\sigma$. Hereafter, we denote the features of $\sigma$ by $x_\sigma$.
\looseness=-1

\noindent \textbf{Topological neural networks (TNNs).} Most TNNs use message-passing mechanisms to obtain cell-level representations \citep{Papillon23}. In particular, let $\mathcal{N}_i$ be a finite sequence of neighborhood structures, $\mathcal{N}_C(\sigma, \tau)=\mathcal{N}_C(\sigma) \cap \mathcal{N}_C(\tau)$, and $\mathcal{N}_B(\sigma, \tau)=\mathcal{N}_B(\sigma) \cap \mathcal{N}_B(\tau)$. In its general form, starting from $h^{0}_\sigma=x_\sigma$ for all $\sigma$, a message-passing TNN \citep{Bodnar2021} recursively computes
\begin{align}
m^{\ell}_{i,\sigma} &= \begin{cases}
\{\!\!\{\phi_{\ell, i}(h_\tau^{\ell}, h_\sigma^{\ell}, h_\delta^{\ell}): \tau \in \mathcal{N}_i(\sigma), \delta \in \mathcal{N}_B(\sigma, \tau)\}\!\!\}, \text{ if } \mathcal{N}_i = 
\mathcal{N}_{\downarrow}\\
\{\!\!\{\phi_{\ell, i}(h_\tau^{\ell}, h_\sigma^{\ell}, h_\delta^{\ell}): \tau \in \mathcal{N}_i(\sigma), \delta \in \mathcal{N}_C(\sigma, \tau)\}\!\!\}, \text{ if } \mathcal{N}_i = \mathcal{N}_{\uparrow}\\
\{\!\!\{\phi_{\ell, i}(h_\tau^{\ell}, h_\sigma^{\ell}): \tau \in \mathcal{N}_i(\sigma)\}\!\!\}, \text{otherwise}.
\end{cases}\\
h_\sigma^{\ell+1} &= \varphi\left(h_\sigma^{\ell}, \bigotimes_{i} \mathrm{Agg}_\ell \left(m^{\ell}_{i,\sigma} \right)\right)
\end{align}
where $h_\sigma^{\ell}$ is the embedding of $\sigma$ at layer $\ell$, $\bigotimes$ and $\mathrm{Agg}_\ell$ are inter- and intra-neighborhood aggregation functions, respectively, and $\varphi$ is an update function (e.g., MLP).

\noindent \textbf{Graph lifting.} A \emph{graph lifting} is a map $\mathrm{lift}: \mathbb{G} \rightarrow \mathbb{T}$ from the space of attributed graphs, $\mathbb{G}$, to a target domain, $\mathbb{T}$, such that $G \cong_{\mathbb{G}} G' \implies \mathrm{lift}(G) \cong_{\mathbb{T}} \mathrm{lift}(G')$, where $\cong_\mathbb{T}$ denotes the \emph{isomorphism} relation in domain $\mathbb{T}$. 
One of the most widely used methods for lifting graphs to cell complexes is \emph{cycle lifting} \citep{Bodnar2021}. This is a static (non-learnable) approach that constructs 2-cells by identifying basic cycles (elements of a cycle basis) or chordless cycles \citep{Bodnar2021} in input graphs. Specifically, the vertices involved in a basic cycle are grouped to form a 2-cell in the resulting cell complex. A cycle basis of a graph $G$ is a minimal set of cycles such that any other cycle in $G$ can be expressed as a modulo-2 sum of cycles from this set.





\section{Differentiable Lifting}
In this section, we introduce $\difflift$ (read DiffLift), a general framework for learning graph lifting functions. 
\autoref{sec:overview} provides an iterative description of our method, allowing for learning structures of increasingly higher order --- when the target domain is hierarchically structured.
\autoref{sec:hypergraph} and \autoref{sec:cell-complex}  instantiate $\difflift$ for graph-to-hypergraph and graph-to-cell-complex liftings, respectively.
Moreover, we formulate our approach for simplicial- and combinatorial complexes in \autoref{ap:extensions}. \looseness=-1

\subsection{General formulation}
\label{sec:overview}

Lifting consists of determining which higher-order cells should be added to an input graph $G$, satisfying the constraints of the target domain. To do so, we propose the following recipe. 
\looseness=-1

\begin{mybox}{$\difflift:$  general recipe for differentiable graph liftings}
\noindent \textbf{Input}: Attributed graph $G=(V, E, x)$, target domain $\mathbb{T}$, and maximum dimension $D_\text{max}$.

\vspace{8pt}
\noindent \textbf{Step 1: Compute node embeddings.} Use an arbitrary GNN to compute a vector representation (embedding) $z_v$ for each node $v \in V$. This GNN component can be either a pre-trained model or learned end-to-end.  Set the current domain dimension to $D = 1$.

\vspace{8pt}
\noindent \textbf{Step 2: Elicit candidate cells.} Given the node embeddings $\{z_v\}_{v \in V}$, define a set of candidate cells $\mathcal{C} \subseteq 2^{V}$ of dimension $D$. For each cell $C \in \mathcal{C}$, compute an embedding $z_C = \bigoplus_{v \in C} z_v$, where $\bigoplus$ is an arbitrary permutation-invariant aggregation function. Note that the exact procedure for defining candidate cells depends on the target domain $\mathbb{T}$, as candidates must respect possible hierarchical constraints. \looseness=-1



\vspace{8pt}
\noindent \textbf{Step 3: Accept/reject candidate cells.} Apply a neural network $\phi$ (e.g., an MLP) that defines an acceptance probability $\phi(z_C)$ for each candidate cell $C$. Finally, draw a sample $y_C$ from a Bernoulli distribution with parameter $\phi(z_C)$ indicating whether cell $C$ is accepted or not. The resulting domain is then given by $V \cup E \cup \{C \in \mathcal{C}: y_C = 1 \text{ with } y_C \sim \text{Ber}(\phi(z_C))\}$.

\vspace{8pt}
\noindent \textbf{Step 4: Termination check.} If $D=D_\text{max}$, halt; otherwise, $D \leftarrow D+1$ and return to Step 2.\looseness=-1
\end{mybox}

Importantly, $\difflift$ is learned in an end-to-end fashion, using the straight-through estimator \citep{Bengio2013} to propagate gradients through samples at Step 3. For hypergraphs, we assume hyperedges have dimension one, causing $\difflift$ to stop once it reaches Step 4. We note that \textcolor{lb}{Steps 2 and 3 are the only domain-dependent ingredients of our algorithm}. Next, we explain how these steps can be adapted to specific domains.
\looseness=-1

\subsection{Graph-to-hypergraph lifting}
\label{sec:hypergraph}

\texttt{\textcolor{lb}{$\Rightarrow$[Step 2]}} For notational convenience, suppose we wish to learn up to one hyperedge per node.  For each node $v$, we define a candidate hyperedge $C(v)$ using the $k_v$ nearest neighbors of $v$ in the embedding space:
\begin{equation}
    C(v)=\{S \subset V: |S|=k_v \text{ and } w \notin S \implies \mathrm{dist}(z_{w}, z_v) \geq \max_{u \in S} \mathrm{dist}(z_u, z_v) \},
\end{equation}
where $\mathrm{dist}(\cdot, \cdot)$ denotes a dissimilarity metric. Here, we consider the Euclidean distance.

To allow for adaptive hyperedge sizes, we sample $k_v$ according to a probability distribution parameterized by (a function of) $v$'s embedding $z_v$. More specifically, we define the ${(k_\text{max} - k_\text{min}+1)}$-dimensional probability vector $\pi_v \propto \exp \circ\, \mathrm{MLP}(z_v)$ and draw $k_v \sim \mathrm{Categorical}(\pi_v)$, where $k_\text{min}$ and $k_\text{max}$ are lower- and upper-bounds on $k_v$.

\texttt{\textcolor{lb}{$\Rightarrow$[Step 3]}} We define the probability of acceptance (i.e., of $b_v=1$) for $C(v)$ as a function of the (multiset of embeddings of) nodes in $C(v)$. More specifically, we define $b_v$ as
\begin{equation}
b_v \sim \mathrm{Ber}(\Psi(\multiset{z_u: u \in C(v)})),
\end{equation}
where $\Psi$ is learned and maps from multisets (i.e., $\Psi$ is order-invariant) of elements in $\mathbb{R}^d$ to $(0,1)$. 

\noindent \textbf{Feature lifting.}  Each accepted hyperedge $C(v)$ receives a feature vector $x_{C(v)}$ computed as a multiset operation over $\multiset{x_u : u \in C(v)}$. Specifically, we employ a \emph{scaled sum projection}:\looseness=-1  
\begin{equation}
    x_{C(v)} = \frac{1}{k_v} \sum_{u \in C(v)} x_u, \quad \forall v \text{ such that } b_v=1.
\end{equation}





\subsection{Graph-to-cell-complex lifting}
\label{sec:cell-complex}


\begin{figure}[t]
    \centering
    \includegraphics[width=\linewidth]{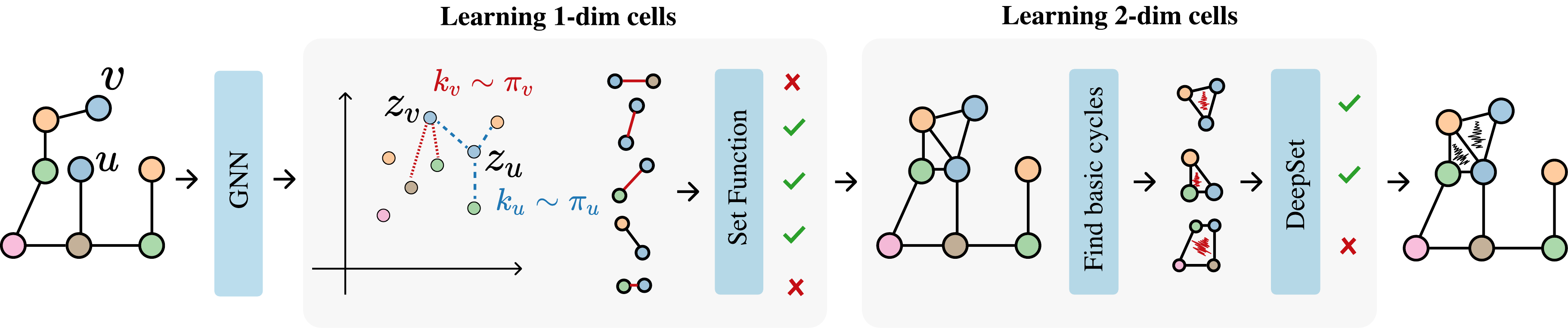}
    \caption{Two iterations of $\difflift$ for cell complexes.
    At the first iteration, we leverage each node $u$'s GNN embedding $z_u$ to delineate candidate 1-dim cells. Specifically, we consider cell-equivalent of edges linking $u$ to each of its $k_u$ NNs in embedding space, where $k_u$ is a random variable parameterized by $z_u$. We use a set function over the embedding of nodes within each cell to compute their acceptance probabilities. At the second iteration onwards, we use cycle lifting in our augmented cell complex to elicit candidate cells, whose acceptance probabilities are computed similarly to the first step. \looseness=-1 
    }
    \label{fig:difflift-cell-complex}
\end{figure}

For computational reasons, we split the lifting procedure for cell complexes into two cases. We provide an overview of our proposed graph-to-cell-complex lifting in \autoref{fig:difflift-cell-complex}.

\noindent \underline{\textbf{Case $D=1$: Learning edges}}  

\texttt{\textcolor{lb}{$\Rightarrow$[Step 2]}} Similarly to Step 2 of graph-to-hypergraph lifting, for each node $v$, we sample a neighborhood size $k_v$ and define a set $C(v) \subseteq V \setminus {\{v\}}$ containing the nodes associated with the $k_v$ nearest neighbors of $v$ in the embedding space, excluding $v$ itself. 

\texttt{\textcolor{lb}{$\Rightarrow$[Step 3]}} Next, we construct candidate edges ($1$-cells) by considering each pair $(v, v^\prime)$ and define their probability of acceptance (i.e., of $b_{v,v^\prime}=1$) as a function of the embeddings of $v$ and $v^\prime$. Specifically, we set $b_{v,v^\prime} \sim \mathrm{Ber}\left(\Psi(\multiset{z_v,\,z_{v^\prime}})\right)$, where $\Psi$ is an order-invariant function. 

At end of this iteration, the obtained cell complex is given by:
\begin{equation}
K^1= V \cup E \cup \{ \{v, v'\} : v \in V, v^\prime \in C(v) \text{ with } b_{v, v^\prime}=1 \}.
\end{equation} 

Regarding \emph{feature lifting}, we apply scaled sum projection, identically to the hypergraph case. 

\noindent \underline{\textbf{Case $D\geq2$: Learning $D$-cells}}  

To select candidate cells of arbitrary dimension, we need the notion of $n$-cycles of a cell complex. Let $C_n(K)$ denote the $n$-chains of the cell complex $K$ equipped with $\mathbb{Z}/2\mathbb{Z}$-vector space structure. Also let $\partial_n: C_n(K) \rightarrow C_{n-1}(K)$ be the boundary linear map on $K$. Then, the $n$-cycles of $K$ are given by $Z_n(K) = \mathrm{ker}(\partial_n)$. We provide further details in the supplementary material.

\texttt{\textcolor{lb}{$\Rightarrow$[Step 2]}} Let $K^{D-1}$ be the cell complex at the end of iteration $D-1$. We select a basis for $(D-1)$-set of cycles $Z_{D-1}(K^{D-1})$ in $K^{D-1}$ to serve as candidate cells. Recall a basis for cycles is a minimal collection of cycles such that any cycle can be written as a modulo-2 sum of cycles in the basis.
We note that we can also employ chordless cycles to define the 2-cells, as in \citep{Bodnar2021}.
\looseness=-1


\texttt{\textcolor{lb}{$\Rightarrow$[Step 3]}} Let $\mathcal{C}$ be the set of candidate $(D-1)$-cycles from Step 2. We define the probability of accepting $C \in \mathcal{C}$ (i.e., setting $b_C=1$) using a DeepSet model \citep{DeepSets} over the multiset of embeddings $\multiset{z_v}_{v \in C}$ of all nodes $v$ in $C$. The output complex at this iteration is then 
\begin{equation}
K^D = K^{D-1} \cup \{C \in \mathcal{C}: b_C=1\}.
\end{equation}
For simplicity, again, the features of $D$-cells are obtained via sum projection lifting.

\begin{remark} Despite the generality of $\difflift$, in the experiments we only consider 2-dimensional cell complexes ($D_{\text{max}}=2$) and use the algorithm in~\citep{cycle_basis_alg} (available at the toolbox NetworkX~\citep{networkx}) to identify basic 1-cycles in graphs. This is mainly due to the fact that current implementations of TNNs for cell complexes only support 2-dimensional objects --- for instance, see TopoBench~\citep{telyatnikov2024topobenchmarkx}.
\end{remark}

\begin{remark} We can obtain a deterministic version of $\difflift$ using a probability threshold, i.e., we simply set $b_C = \mathbbm{1}[\Psi(\cdot) > \gamma]$ with, e.g., $\gamma=0.5$, for all candidate cells $C$. 
\end{remark}

Experiments regarding the deterministic version can be found in the Appendix \ref{ap:deterministic}.

\section{Related Works}
\paragraph{Topological deep learning.}
Traditional graph deep learning methods are limited to modeling only pairwise interactions, making them unsuitable for capturing higher-order dependencies involving multiple nodes \citep{hajij2022topological, papillon2023icml}. To address this limitation, a variety of deep topological learning methods have been developed for hypergraphs \citep{bai2021hypergraph, yadati2019hypergcn}, simplicial complexes \citep{hajij2021simplicial, goh2022simplicial, maggs2023simplicial, yang2022simplicial}, and cell complexes \citep{hajij2022topological, hajij2020cell}, the works that are based on the topological signal processing field~\citep{barbarossa2020topological, schaub2021signal, roddenberry2022signal, sardellitti2021topological}. \citet{papillon2025topotuneframeworkgeneralized} also use GNNs to enhance TDL, where the lifted topological domain is transformed into augmented Hasse graphs. These methods have demonstrated their effectiveness across several practical applications, including action recognition \citep{wang2022survey, hao2021hypergraph}, bioinformatics \citep{liu2022multi}, and neuroscience \citep{wang2023dynamic}.

\paragraph{Liftings to topological domains.}
Most relational datasets and benchmarks are defined on discrete structures such as graphs. To apply topological deep learning methods to these datasets, a transformation process known as lifting is required, which maps discrete data into topological domains \citep{telyatnikov2024topobenchmarkx, hajij2022topological, bernardez2024icml}. This lifting process can be either predefined -- e.g., based on structural features like node proximity or the presence of cycles -- or learned directly from the data \citep{battiloro2023latent, kazi2022differentiable}. Graph structure learning methods \citep{qian2024probabilistic, kazi2022differentiable, franceschi2019learning, topping2021understanding, sun2023self, chen2020iterative, jin2020graph} are closely related to the graph lifting literature and can be interpreted as instances of graph lifting to graph domain. Our approach represents the most general form of learnable lifting proposed so far and empirically outperforms the aforementioned methods in many benchmarks.
\looseness=-1


\paragraph{Static liftings.}
To the best of our knowledge, \citet{bodnar2021weisfeiler,Bodnar2021} were the first to combine static liftings and high-order message passing, focusing on simplicial- and cell complexes. These static liftings embed a graph into a topological domain by, e.g., aggregating each node’s n-hop neighborhood or by tracing its cycles. The repertoire of static liftings was later broadened by the ICML TDL challenge~\citep{bernardez2024icml}, which added methods based on kNN, Voronoi decompositions, and random walks. Our work proposes a more flexible, data-driven approach to defining liftings, which offers benefits across a range of tasks.


\section{Experiments}
In this section, we evaluate $\difflift$ on two complementary tasks: graph classification and node classification. We compare it against broadly used lifting schemes for both hypergraphs and cell complexes. We also report results across different TNNs within each of these domains. We run experiments using PyTorch~\citep{pytorch} and PyTorch Geometric~\citep{Fey/Lenssen/2019}; our code is anonymously available at \url{https://github.com/JorgeLuizFranco/difflifting}.

\subsection{Graph classification}
\label{section:graph_classification}

\textbf{Datasets.} We evaluate model performance on six widely used graph-level benchmark datasets for molecular property prediction: NCI1, NCI109, MUTAG, MOLHIV, PROTEINS, and ZINC~\citep{TUDatasets, dwivedi2023benchmarking, hu2020ogb}. These datasets are standard benchmarks in the literature for assessing the effectiveness of graph-based models \citep{dwivedi2023benchmarking,telyatnikov2024topobenchmarkx}. All tasks are binary classification problems, with the exception of ZINC, which is a regression task. We provide more details regarding datasets in the Appendix.
\looseness=-1

\textbf{Baselines.} We compare $\difflift$ with four existing graph lifting methods: cycle lifting, $k$-hop lifting, $k$-nearest-neighbor ($k$-NN) lifting, and kernel lifting. Among these, cycle lifting is the most widely adopted strategy for graph-to-cell-complex liftings and has become the \emph{de facto} standard in most of TNNs operating on cell complexes \citep{telyatnikov2024topobenchmarkx}. Similarly, $k$-hop lifting is the predominant approach for constructing hypergraphs from graphs and is often the sole method considered in recent benchmarks such as \citep{telyatnikov2024topobenchmarkx}. We also consider $k$-NN lifting as it shares similarities with our approach due to the use of $k$-NN. Finally, we consider kernel lifting, one of the most successful approaches in the ICML TDL challenge~\citep{bernardez2024icml}. Notably, our choice of baselines covers lifting methods based on connectivity (cycle and $k$-hop liftings), features ($k$-NN lifting), and both connectivity and features (kernel lifting).
We provide formulations for the baseline liftings in \autoref{ap:liftings}. We consider the following TNNs: CWN, CIN \citep{bodnar2021weisfeiler} and CXN \citep{hajij2020cell} for cell complexes, and UniGCNII and UniGIN \citep{huang2021unignn} for hypergraphs. \looseness=-1

\textbf{Evaluation setup.} For ZINC and MOLHIV, we use the publicly available train/val/test data splits; for the remaining datasets, we use a random 80/10/10\% split. We optimize the hyper-parameters of the lifting methods and take the optimal hyperparameter values from \citep{telyatnikov2024topobenchmarkx} whenever available; otherwise, we select optimal values based on the optimal results using cycle or $k$-hop lifting. We provide further details on the choice of hyperparameters and model selection in the supplementary material. 
We compute the mean and standard deviation of the performance metrics (MAE \(\downarrow\) for ZINC, AUC \(\uparrow\) for MOLHIV, and accuracy \(\uparrow\) for all other datasets) over three independent runs.
\looseness=-1

\begin{table}[htb]
    \centering
    \caption{Graph classification: $\difflift$ vs static liftings. We denote the best-performing model for each dataset/TNN in bold. For any fixed TNN and dataset, $\difflift$ is better than static liftings in $90$\% of cases, offering a performance improvement of up to $45$\%. }
    \vspace{2pt}
    \resizebox{\textwidth}{!}{
    \begin{tabular}{lllcccccc}
        \toprule
        \textbf{Domain} & \textbf{TNN} &  \textbf{Lifting} &  \textbf{NCI1}$\uparrow$ &  \textbf{NCI109} $\uparrow$ & \textbf{MOLHIV}$\uparrow$ & 
        \textbf{MUTAG}$\uparrow$ & \textbf{Proteins}$\uparrow$
        & \textbf{ZINC}$\downarrow$ \\
        \midrule
        \multirow{2}{*}{Graph} &  \multirow{1}{*}{GCN} & - & $74.45\textcolor{gray}{\scriptstyle{\pm1.05}}$    & $76.46\textcolor{gray}{\scriptstyle{\pm1.03}}$ & $74.99\textcolor{gray}{\scriptstyle{\pm1.09}}$ & $64.91\textcolor{gray}{\scriptstyle{\pm 4.96}}$ & $70.18\textcolor{gray}{\scriptstyle{\pm1.35}}$ & $0.64 \textcolor{gray}{\scriptstyle{\pm0.04}}$ \\
        &  \multirow{1}{*}{GIN} & - & $76.89\textcolor{gray}{\scriptstyle{\pm 1.75}}$    & $76.90\textcolor{gray}{\scriptstyle{\pm 0.80}}$ & $70.76\textcolor{gray}{\scriptstyle{\pm 2.46}}$ & $80.70\textcolor{gray}{\scriptstyle{\pm 2.48}}$ & $72.50\textcolor{gray}{\scriptstyle{\pm 2.31}}$ & $0.59 \textcolor{gray}{\scriptstyle{\pm0.03}}$\\
        \cmidrule{1-9}
        
        \multirow{8}{*}{Cellular} &  \multirow{2}{*}{CWN} & Cycle & $76.93 \textcolor{gray}{\scriptstyle{\pm 1.18}}$    & $76.71 \textcolor{gray}{\scriptstyle{\pm 1.34}}$ & $70.15 \textcolor{gray}{\scriptstyle{\pm 3.98}}$ & $66.67 \textcolor{gray}{\scriptstyle{\pm 12.41}}$ &  $69.05 \textcolor{gray}{\scriptstyle{\pm 2.95}}$ & $0.46 \textcolor{gray}{\scriptstyle{\pm 0.01}}$ \\
         
         & & \textcolor{lb}{$\mathrm{\partial lift}$} &  $\textbf{79.81} \textcolor{gray}{\scriptstyle{\pm 0.40}}$  &  $\textbf{80.55} \textcolor{gray}{\scriptstyle{\pm 0.50}}$  & $\textbf{75.37} \textcolor{gray}{\scriptstyle{\pm 0.80}}$ & $\textbf{85.96} \textcolor{gray}{\scriptstyle{\pm 4.96}}$ &  $\textbf{70.54} \textcolor{gray}{\scriptstyle{\pm 3.34}}$ & $\textbf{0.17}\textcolor{gray}{\scriptstyle{\pm 0.00}}$  \\
         
        \cmidrule{2-9}
         & \multirow{2}{*}{CXN} & Cycle & $72.02 \textcolor{gray}{\scriptstyle{\pm 1.69}}$  & $75.01 \textcolor{gray}{\scriptstyle{\pm 0.62}}$ & $69.17 \textcolor{gray}{\scriptstyle{\pm 1.20}}$ & $61.40 \textcolor{gray}{\scriptstyle{\pm 2.48}}$ & $\textbf{70.83} \textcolor{gray}{\scriptstyle{\pm 1.52}}$ & $0.79 \textcolor{gray}{\scriptstyle{\pm 0.02}}$     \\
         & & \textcolor{lb}{$\mathrm{\partial lift}$} & $\textbf{82.08} \textcolor{gray}{\scriptstyle{\pm 1.50}}$  & $\textbf{82.57} \textcolor{gray}{\scriptstyle{\pm 0.40}}$ &  $\textbf{74.83} \textcolor{gray}{\scriptstyle{\pm 1.96}}$ & $\textbf{84.21} \textcolor{gray}{\scriptstyle{\pm 4.30}}$ & $69.94 \textcolor{gray}{\scriptstyle{\pm 2.10}}$ & $\textbf{0.17} \textcolor{gray}{\scriptstyle{\pm 0.01}}$   \\
        
        \cmidrule{2-9}
         & \multirow{2}{*}{CIN} & Cycle & $75.91 \textcolor{gray}{\scriptstyle{\pm 1.11}}$ & $76.11 \textcolor{gray}{\scriptstyle{\pm 1.09}}$ &  $68.46 \textcolor{gray}{\scriptstyle{\pm 2.16}}$ & $66.96 \textcolor{gray}{\scriptstyle{\pm 1.46}}$ & $67.86 \textcolor{gray}{\scriptstyle{\pm 0.89}}$ &   $0.42 \textcolor{gray}{\scriptstyle{\pm 0.01}}$  \\
         & & \textcolor{lb}{$\mathrm{\partial lift}$} & $\textbf{79.59} \textcolor{gray}{\scriptstyle{\pm 1.50}}$  & $\textbf{81.06} \textcolor{gray}{\scriptstyle{\pm 0.40}}$ &  $\textbf{72.37} \textcolor{gray}{\scriptstyle{\pm 1.65}}$ & $\textbf{88.72} \textcolor{gray}{\scriptstyle{\pm 4.30}}$ & $\textbf{72.43} \textcolor{gray}{\scriptstyle{\pm 2.10}}$ & $\textbf{0.20} \textcolor{gray}{\scriptstyle{\pm 0.01}}$   \\
        
        \cmidrule{1-9}

        \multirow{10}{*}{Hypergraph} & \multirow{4}{*}{UniGCN2} & $k$-hop & $72.70 \textcolor{gray}{\scriptstyle{\pm 0.52}}$ & $72.01 \textcolor{gray}{\scriptstyle{\pm 1.55}}$ & $50.72 \textcolor{gray}{\scriptstyle{\pm 1.06}}$ & $61.40 \textcolor{gray}{\scriptstyle{\pm 2.48}}$  & $72.92 \textcolor{gray}{\scriptstyle{\pm 1.11}}$ & $0.66 \textcolor{gray}{\scriptstyle{\pm 0.02}}$   \\
        ~ &  & $k$-NN & 
    71.78$\textcolor{gray}{\scriptstyle{\pm 0.20}}$ & 
    $68.60\textcolor{gray}{\scriptstyle{\pm 0.93}}$ & 
    $57.73\textcolor{gray}{\scriptstyle{\pm 6.84}}$ & 
    $64.91\textcolor{gray}{\scriptstyle{\pm 2.48}}$ & 
    $73.51\textcolor{gray}{\scriptstyle{\pm 0.42}}$ & $1.10\textcolor{gray}{\scriptstyle{\pm 0.01}}$  
      \\
        ~ &  & kernel & $73.80 \textcolor{gray}{\scriptstyle{\pm 0.94}}$   & $72.64 \textcolor{gray}{\scriptstyle{\pm 0.40}}$ & $57.07 \textcolor{gray}{\scriptstyle{\pm 10.32}}$ & $63.16 \textcolor{gray}{\scriptstyle{\pm 8.59}}$ &  $73.21 \textcolor{gray}{\scriptstyle{\pm 0.73}}$ & $0.79 \textcolor{gray}{\scriptstyle{\pm 0.02}}$ \\
        ~ &  & \textcolor{lb}{$\mathrm{\partial lift}$} & $\textbf{77.45} \textcolor{gray}{\scriptstyle{\pm 1.88}}$   & $\textbf{75.30} \textcolor{gray}{\scriptstyle{\pm 1.10}}$ & $\textbf{69.32} \textcolor{gray}{\scriptstyle{\pm 1.62}}$ & $\textbf{89.47} \textcolor{gray}{\scriptstyle{\pm 4.30}}$ &  $\textbf{73.51} \textcolor{gray}{\scriptstyle{\pm 0.84}}$ & $\textbf{0.56} \textcolor{gray}{\scriptstyle{\pm 0.03}}$ \\
        
        \cmidrule{2-9}
        & \multirow{4}{*}{UniGIN} & $k$-hop & $65.50 \textcolor{gray}{\scriptstyle{\pm 1.99}}$  & $66.97 \textcolor{gray}{\scriptstyle{\pm 7.25}}$ & $63.49 \textcolor{gray}{\scriptstyle{\pm 9.55}}$ & $64.91 \textcolor{gray}{\scriptstyle{\pm 2.48}}$  & $71.43 \textcolor{gray}{\scriptstyle{\pm 0.73}}$  & $1.15 \textcolor{gray}{\scriptstyle{\pm 0.01}}$   \\
        ~ &  & $k$-NN & 
    $\textbf{72.83}\textcolor{gray}{\scriptstyle{\pm 1.09}}$ & 
    $70.14\textcolor{gray}{\scriptstyle{\pm 1.48}}$ & 
    $52.34\textcolor{gray}{\scriptstyle{\pm 3.21}}$ & 
    $59.65\textcolor{gray}{\scriptstyle{\pm 4.96}}$ & 
    $72.62\textcolor{gray}{\scriptstyle{\pm 1.52}}$ & $1.10\textcolor{gray}{\scriptstyle{\pm 0.02}}$ \\
        ~ &  & kernel & $60.50 \textcolor{gray}{\scriptstyle{\pm 1.26}}$   & $66.59 \textcolor{gray}{\scriptstyle{\pm 1.49}}$ & $49.60 \textcolor{gray}{\scriptstyle{\pm 0.07}}$ & $57.89 \textcolor{gray}{\scriptstyle{\pm 4.30}}$ &  $66.67 \textcolor{gray}{\scriptstyle{\pm 1.83}}$ & $1.45 \textcolor{gray}{\scriptstyle{\pm 0.02}}$ \\
        ~ &  & \textcolor{lb}{$\mathrm{\partial lift}$} & $64.88 \textcolor{gray}{\scriptstyle{\pm 1.09}}$   & $\textbf{79.74} \textcolor{gray}{\scriptstyle{\pm 0.23}}$ & $\textbf{72.04} \textcolor{gray}{\scriptstyle{\pm 0.88}}$ & $\textbf{66.67} \textcolor{gray}{\scriptstyle{\pm 6.56}}$ &  $\textbf{73.81} \textcolor{gray}{\scriptstyle{\pm 1.52}}$ & $\textbf{0.92} \textcolor{gray}{\scriptstyle{\pm 0.05}}$ \\
        \bottomrule
    \end{tabular}
}
    \label{tab:graph_classific}
\end{table}

\textbf{Results.}
\autoref{tab:graph_classific} shows that $\difflift$ is the best-performing lifting method in over 90\% of the TNN/dataset combinations, both for cell complexes and hypergraphs. 
Notably, $\difflift$ resulted in an improvement in average accuracy of up to $\approx 45$\% compared to static liftings using the same TNN.
For CWN, CIN, and UniGCNII, our method outperforms static liftings on all datasets.
On NCI109 and ZINC, $\difflift$ is consistently better than the static liftings across all TNN backbones.
%
We also observe that TNNs perform better than GNNs in this setup with $\difflift$.
Overall, these results validate the effectiveness of $\difflift$.
\looseness=-1


\textbf{Impact of GNN choice on performance.} We aim to assess how sensitive our approach is to the choice of GNN. We report results using GIN \citep{xu2018how} and GPS \citep{gps_cite}.
\looseness=-1

\autoref{tab:ablation} indicates that choosing GNNs that are able to generate richer and more informative latent node representations leads to better results in $\difflift$. In particular, GPS performs better than GIN in most datasets. A possible explanation for this observation is the greater expressivity of GPS,  which benefits from the incorporation of positional encodings. Notably, on cell complexes and ZINC dataset, GPS allows reducing the MAE from 0.46 to 0.17.

\begin{table}[htb]
    \vspace{-4pt}
    
    \centering
    \caption{Effect of GNN backbone on the performance of $\difflift$. The results suggest that the expressive power of backbone GNNs have a direct impact in $\difflift$'s performance. Except for MOLHIV and NCI1, GPS leads to better performance than GIN overall. \looseness=-1}
    
    \resizebox{\textwidth}{!}{
    \begin{tabular}{llcclccc}
        \toprule
         \textbf{TNN} &  \textbf{GNN} &  \textbf{NCI1}$\uparrow$ &  \textbf{NCI109} $\uparrow$  &\textbf{MOLHIV} $\uparrow$& 
        \textbf{MUTAG}$\uparrow$ & \textbf{Proteins}$\uparrow$
        & \textbf{ZINC}$\downarrow$ \\
        \midrule
          \multirow{2}{*}{CWN} & GPS & $79.81 \textcolor{gray}{\scriptstyle{\pm 0.40}}$  &  $\textbf{80.55} \textcolor{gray}{\scriptstyle{\pm 0.50}}$   &$64.31 \textcolor{gray}{\scriptstyle{\pm 5.32}}$ & $\textbf{87.72} \textcolor{gray}{\scriptstyle{\pm 2.48}}$ &  $70.54 \textcolor{gray}{\scriptstyle{\pm 3.34}}$ & $\textbf{0.17} \textcolor{gray}{\scriptstyle{\pm 0.00}}$  \\
         
         & \color{black}GIN\color{black} &  $\textbf{81.59} \textcolor{gray}{\scriptstyle{\pm 0.80}}$  &  $78.69 \textcolor{gray}{\scriptstyle{\pm 1.43}}$  &$\textbf{75.37} \textcolor{gray}{\scriptstyle{\pm 0.80}}$& $82.46 \textcolor{gray}{\scriptstyle{\pm 2.48}}$ &  $\textbf{71.13} \textcolor{gray}{\scriptstyle{\pm 2.76}}$ & $0.46 \textcolor{gray}{\scriptstyle{\pm 0.00}}$  \\
         
        \cmidrule{1-8}
        \multirow{2}{*}{CXN} & GPS & $79.97 \textcolor{gray}{\scriptstyle{\pm 0.61}}$  & $\textbf{82.57} \textcolor{gray}{\scriptstyle{\pm 0.40}}$   &$65.58 \textcolor{gray}{\scriptstyle{\pm 4.42}}$&  $\textbf{84.21} \textcolor{gray}{\scriptstyle{\pm 4.30}}$ & $\textbf{69.94} \textcolor{gray}{\scriptstyle{\pm 2.10}}$  & $\textbf{0.17} \textcolor{gray}{\scriptstyle{\pm 0.01}}$         \\
         & \color{black}GIN\color{black} & $\textbf{81.35} \textcolor{gray}{\scriptstyle{\pm 2.29}}$  & $79.98 \textcolor{gray}{\scriptstyle{\pm 1.01}}$  &$\textbf{72.25} \textcolor{gray}{\scriptstyle{\pm 3.23}}$& $77.19 \textcolor{gray}{\scriptstyle{\pm 2.48}}$ & $67.86 \textcolor{gray}{\scriptstyle{\pm 2.19}}$ & $0.43 \textcolor{gray}{\scriptstyle{\pm 0.01}}$   \\
        \cmidrule{1-8}

         \multirow{2}{*}{UniGCN2}  & GPS & $\textbf{78.67} \textcolor{gray}{\scriptstyle{\pm 1.46}}$   & $74.50 \textcolor{gray}{\scriptstyle{\pm 1.16}}$   &$68.22 \textcolor{gray}{\scriptstyle{\pm 2.38}}$& $\textbf{89.47} \textcolor{gray}{\scriptstyle{\pm 4.30}}$   & $\textbf{73.81} \textcolor{gray}{\scriptstyle{\pm 0.42}}$   & $\textbf{0.56} \textcolor{gray}{\scriptstyle{\pm 0.03}}$ \\
          & \color{black}GIN\color{black} & $75.38 \textcolor{gray}{\scriptstyle{\pm 1.39}}$   & $\textbf{74.98} \textcolor{gray}{\scriptstyle{\pm 1.12}}$  &$\textbf{68.73} \textcolor{gray}{\scriptstyle{\pm 2.05}}$& $64.91 \textcolor{gray}{\scriptstyle{\pm 6.56}}$ &  $73.51 \textcolor{gray}{\scriptstyle{\pm 0.84}}$ & $0.63 \textcolor{gray}{\scriptstyle{\pm 0.01}}$ \\
        
        \cmidrule{1-8}
        \multirow{2}{*}{UniGIN} & GPS & $\textbf{66.42} \textcolor{gray}{\scriptstyle{\pm 1.79}}$    & $\textbf{79.74} \textcolor{gray}{\scriptstyle{\pm 0.23}}$   &$68.32 \textcolor{gray}{\scriptstyle{\pm 3.12}}$ & $66.67 \textcolor{gray}{\scriptstyle{\pm 6.56}}$ &  $\textbf{72.32} \textcolor{gray}{\scriptstyle{\pm 0.73}}$ & $\textbf{1.01} \textcolor{gray}{\scriptstyle{\pm 0.05}}$  \\
          & \color{black}GIN\color{black} & $64.40 \textcolor{gray}{\scriptstyle{\pm 0.41}}$   & $78.53 \textcolor{gray}{\scriptstyle{\pm 0.69}}$  &$\textbf{68.86} \textcolor{gray}{\scriptstyle{\pm 3.05}}$& $\textbf{70.18} \textcolor{gray}{\scriptstyle{\pm 6.56}}$ &  $72.02 \textcolor{gray}{\scriptstyle{\pm 3.74}}$ & $1.12 \textcolor{gray}{\scriptstyle{\pm 0.26}}$ \\
        
        \bottomrule
    \end{tabular}
}
    \label{tab:ablation}
    \vspace{15pt}
\end{table}


\vspace{10pt}

\subsection{Node classification}

\textbf{Datasets.} For node classification, we evaluate $\difflift$ on four datasets: Cora, Citeseer \citep{yang2016revisitingsemisup}, Texas, and Wisconsin \citep{rozemberczki2020multiscale}. Within these datasets, two are knowingly homophilic (Cora and Citeseer) and two are heterophilic datasets (Texas and Wisconsin).
Dataset statistics can be found in \autoref{ap:datasets}.

\textbf{Baselines.} We also compare our method (for cell domains) against the learnable approach in \citep{battiloro2023latent}, called Differentiable Cell Complex Module (DCM), which was originally evaluated on node classification tasks. To do so, we consider $\difflift$ combined with CWN and TopoTune \citep{papillon2025topotuneframeworkgeneralized}. We also include results of them with cycle lifting. We consider the same hypergraph TNN baselines as in \autoref{section:graph_classification}.
\looseness=-1

\textbf{Evaluation setup.} For all datasets, we use random train/val/test data split with 60/20/20\% split. 
Similarly to the experiments for graph classification,
we optimize the hyper-parameters of the lifting methods and take the optimal TNN hyperparameters from \citet{telyatnikov2024topobenchmarkx} when available. Otherwise, we choose them to maximize the validation accuracy using $k$-hop lifting. For more details, please refer to the supplementary material. 
We report the average accuracy and standard deviation over three independent runs. \looseness=-1

\textbf{Results.} \autoref{tab:cellular_node} compares $\difflift$ against DCM and cycle lifting. Notably, $\difflift$ is the best-performing method in all datasets except for Wisconsin, in which it achieves the second-best performance. It is also worth mentioning that $\difflift$ (with either CWN or TopoTune) outperforms DCM for all datasets, sometimes by a large margin --- c.f., Texas and Cora.


\begin{table}[htb]
    \centering
    \caption{Comparison of DCM and $\difflift$ on node classification. $\difflift$ achieves the highest average performance across all datasets (rhs) and significantly outperforms DCM on heterophilic datasets. \looseness=-1}
      \vspace{2pt}  \resizebox{0.95\textwidth}{!}{
    \begin{tabular}{llcccc|c}
        \toprule
        \textbf{TNN} &  \textbf{Lifting} &  \textbf{Cora} &  \textbf{Citeseer} & \textbf{Texas} & \textbf{Wisconsin} & \textbf{Avg } \\
        \midrule
        GCN & - & $85.64\pm0.51$ & $70.43\pm0.71$ & $58.91\pm0.76$ & $49.14\pm0.66$ & $66.03$ \\
        GAT & - & $86.17\pm0.33$ & $73.82\pm0.45$ & $58.38\pm1.05$ & $49.41\pm0.95$ & $66.95$ \\
        GIN & - & $85.50\pm0.54$ & $72.20\pm0.60$ & $59.10\pm0.80$ & $48.50\pm0.70$ & $66.33$ \\
        \midrule
        DCM & -  & $80.73\pm0.33$ & $77.90\pm0.80$ & $56.76\pm6.62$ & $73.86\pm1.85$ & $72.31$ \\
        CWN & Cycle         & $74.80\pm0.08$ & $75.83\pm0.90$ & $63.06\pm7.75$ & $\mathbf{80.39}\pm4.24$ & $73.52$  \\
        CWN & $\partial$Lift & $80.17\pm1.59$ & $72.83\pm2.15$ & $\mathbf{80.18}\pm3.37$ & $77.78\pm3.70$ & $77.74$  \\
        TopoTune & Cycle & $69.03 \pm 0.88$  & $72.90 \pm 0.85$ & $71.56\pm1.27$ & $70.16 \pm 1.85$  & $70.91$  \\
        TopoTune & $\partial$Lift & $\mathbf{86.82}\pm0.75$ & $\mathbf{78.23}\pm1.08$ & $72.97\pm0.00$ & $65.36\pm2.45$ & $\textbf{75.84}$  \\
        \bottomrule
    \end{tabular}
}
    \label{tab:cellular_node}
\end{table}

\autoref{tab:hypergraph_node} reports results of lifting methods for hypergraph neural networks. Compared to $k$-hop, our approach is better on heterophilic datasets but worse on homophilic ones for UniGCN2. 
Additionally, \autoref{tab:hypergraph_node} shows that $\difflift$ leads to better average accuracy than other static liftings.

\begin{table}[htb]
    \centering
    \caption{Comparison of $\difflift$ and static lifting baselines for hypergraphs on node classification. Our method outperforms all static liftings on average across the selected node classification datasets.}
\resizebox{0.95\textwidth}{!}{
    \begin{tabular}{llcccc|c}
        \toprule
        \textbf{TNN} &  \textbf{Lifting} &  \textbf{Cora} &  \textbf{Citeseer} & \textbf{Texas} & \textbf{Wisconsin} & \textbf{Avg} \\
        \midrule

         \multirow{4}{*}{UniGCN2} 
            & $k$-hop        & $\textbf{86.03}\pm0.63$ & $\textbf{78.40}\pm0.36$ & $66.67\pm6.74$ & $69.28\pm5.62$ &  $75.09$ \\
            & $k$-NN         & $74.00\pm0.65$          & $18.17\pm0.05$          & $\textbf{70.27}\pm3.82$ & $\textbf{79.74}\pm2.45$ & $60.54$ \\
            & kernel         & $29.93\pm1.33$          & $18.07\pm0.21$          & $57.66\pm7.75$           & $57.52\pm10.42$         & $40.79$ \\
            & $\partial$Lift & $81.93\pm1.11$          & $78.03\pm0.91$          & $69.37\pm2.55$           & $73.20\pm5.62$          & $\textbf{75.63}$  \\
        \cmidrule{1-7}
        \multirow{4}{*}{UniGIN} 
            & $k$-hop        & $78.73\pm0.66$          & $74.47\pm1.72$          & $65.77\pm9.19$           & $58.82\pm5.77$          & $69.44$ \\
            & $k$-NN         & $62.00\pm1.08$          & $19.33\pm0.48$          & $65.77\pm1.27$           & $\textbf{73.20}\pm4.03$ & $55.07$  \\
            & kernel         & $40.93\pm2.52$          & $18.53\pm0.61$          & $58.56\pm7.09$           & $51.63\pm4.03$          & $42.41$ \\
            & $\partial$Lift & $\textbf{84.23}\pm0.53$ & $\textbf{77.97}\pm0.45$ & $63.96\pm6.37$           & $63.40\pm4.03$          & $\textbf{72.39}$ \\
        \bottomrule
    \end{tabular}
}

    \label{tab:hypergraph_node}
\end{table}



\section{Conclusion}

Topological neural networks (TNNs) are receiving increasing attention in the graph machine learning community. Yet, their effectiveness depends crucially on the choice of graph lifting procedure. Despite its central role, lifting has remained largely unsupervised and task-agnostic, which can lead to the construction of suboptimal topological representations for downstream learning.

To address this limitation, we introduced $\difflift$, a general-purpose, differentiable lifting framework that is compatible with multiple topological domains. Across a broad set of benchmarks and TNN architectures, $\difflift$ consistently outperformed traditional unsupervised lifting methods, demonstrating the benefit of making the lifting process learnable and task-informed.

\looseness=-1

\textbf{Limitations.} For hypergraph domains, $\difflift$ can create candidate hyper-edges and decide whether to keep them in an embarrassingly parallel fashion --- rendering $\difflift$ especially compute-efficient for this domain. However, for cell complexes, we need to compute a cycle basis to elicit candidate cells, which may come at a cubic cost with respect to the number of nodes in the input graph. In this case, we may reduce the number of candidate cells by, for instance, regularizing the $k_v$ variables or shifting their distribution towards zero. Nonetheless, devising more efficient algorithms for candidate identification in hierarchical domains is a clear direction of improvement for future works. 

\textbf{Future work.} Our method can be extended to other topological domains, such as point clouds, making it applicable to 3D mapping tasks. Additionally, future work could focus on addressing the computational challenges of DiffLift and scaling it to handle larger graphs.
Another promising direction is to explore differentiable lifting in dynamic or temporal graphs, where topological structures evolve over time. Moreover, integrating $\difflift$ with pretraining strategies could yield generalizable topological priors across tasks.\looseness=-1

We also believe that formally analyzing the impact of enriching topological structures with learnable liftings on mitigating oversmoothing and oversquashing in TNNs is an interesting research direction. For instance, in scenarios where long and narrow paths connect dense substructures, static liftings are limited to adding cycles within each local region, leaving information between communities to traverse the original bottleneck. In contrast, $\difflift$ can learn to introduce 2-cells that effectively create shortcuts across such bridges, reducing the effective distance between distant nodes and improving information flow. While a full theoretical treatment is left for future work, this adaptive ability provides a plausible mechanism for mitigating oversquashing.

%

\textbf{Ethics Statement.}
We do not identify any immediate, direct societal harms from the technical contributions presented in this work. Our method operates on standard, non-sensitive benchmarks and does not require or expose personally identifiable information.

\textbf{Reproducibility Statement.} The repository containing the code is available at \url{https://github.com/JorgeLuizFranco/difflifting}. We provide further details on used datasets and implementation details (e.g., parameter selection) in Appendices \ref{ap:datasets} and \ref{ap:implementation_details}. \looseness=-1

\section*{Acknowledgments} 

We acknowledge the support by the Coordenação de Aperfeiçoamento de Pessoal de Nível Superior (CAPES) (88887.176396/2025-00),  Fundação Carlos Chagas Filho de Amparo à Pesquisa do Estado do Rio de Janeiro (FAPERJ) (SEI-260003/020348/2025, SEI-260003/020694/2025) and the Conselho Nacional de Desenvolvimento Científico e Tecnológico (CNPq) (404336/2023-0, 305692/2025-9,  315158/2023-9, 163868/2023-9, 408974/2025-7, 312068/2025-5). We also acknowledge the computational resources provided by the Aalto Science-IT Project.




{
\small
\bibliographystyle{plainnat} \bibliography{references}

@inproceedings{xu2018how,
	title        = {How Powerful are Graph Neural Networks?},
	author       = {Keyulu Xu and Weihua Hu and Jure Leskovec and Stefanie Jegelka},
	year         = 2019,
	booktitle    = {International Conference on Learning Representations},
	url          = {https://openreview.net/forum?id=ryGs6iA5Km}
}

@inproceedings{gps_cite,
author = {Ramp\'{a}\v{s}ek, Ladislav and Galkin, Mikhail and Dwivedi, Vijay Prakash and Luu, Anh Tuan and Wolf, Guy and Beaini, Dominique},
title = {Recipe for a general, powerful, scalable graph transformer},
year = {2022},
isbn = {9781713871088},
publisher = {Curran Associates Inc.},
address = {Red Hook, NY, USA},
booktitle = {Proceedings of the 36th International Conference on Neural Information Processing Systems},
articleno = {1054},
numpages = {15},
location = {New Orleans, LA, USA},
series = {NIPS '22}
}

@book{Hatcher02,
  author = {Hatcher, Allen},
  publisher = {Cambridge University Press},
  title = {Algebraic topology},
  year = 2002
}

@article{Hansen2019,
    author = {J. Hansen and R. Ghrist},
    title = {Toward a spectral theory of cellular sheaves},
    journal = {Journal of Applied and Computational Topology},
    year = 2019
}

@article{Bengio2013,
	author = {Yoshua Bengio and Nicholas Léonard and Aaron Courville},
	journal = {ArXiv e-prints},
	title = {Estimating or Propagating Gradients Through Stochastic Neurons for Conditional Computation},
	year = {2013},
}

@InProceedings{Verma2024,
  title = 	 {Topological Neural Networks go Persistent, Equivariant, and Continuous},
  author =       {Verma, Yogesh and Souza, Amauri H and Garg, Vikas},
  booktitle = 	 {International Conference on Machine Learning (ICML)},
  year = 	 {2024},
}

@article{scarselli2009

,
author = {F. Scarselli  and M. Gori and A. C. Tsoi and M. Hagenbuchner and G. Monfardini}, 
title = {The Graph Neural Network Model}, 
year = {2009}, 
publisher = {IEEE Press}, 
volume = {20}, 
number = {1}, 
journal = {IEEE Transactions on Neural Networks},
pages = {61–80}, 
}

@InProceedings{Gilmer2017,
  title = 	 {Neural Message Passing for Quantum Chemistry},
  author =       {J. Gilmer and S. S. Schoenholz and P. F. Riley and O. Vinyals and G. E. Dahl},
  booktitle = 	 {International Conference on Machine Learning (ICML)},
  year = 	 {2017},
}

@InProceedings{Papamarkou2024,
  title = 	 {Position: Topological Deep Learning is the New Frontier for Relational Learning},
  author =       {Papamarkou, Theodore and Birdal, Tolga and Bronstein, Michael M. and Carlsson, Gunnar E. and Curry, Justin and Gao, Yue and Hajij, Mustafa and Kwitt, Roland and Lio, Pietro and Di Lorenzo, Paolo and Maroulas, Vasileios and Miolane, Nina and Nasrin, Farzana and Natesan Ramamurthy, Karthikeyan and Rieck, Bastian and Scardapane, Simone and Schaub, Michael T and Veli\v{c}kovi\'{c}, Petar and Wang, Bei and Wang, Yusu and Wei, Guowei and Zamzmi, Ghada},
  booktitle = 	 {Proceedings of the 41st International Conference on Machine Learning},
  pages = 	 {39529--39555},
  year = 	 {2024},
  editor = 	 {Salakhutdinov, Ruslan and Kolter, Zico and Heller, Katherine and Weller, Adrian and Oliver, Nuria and Scarlett, Jonathan and Berkenkamp, Felix},
  volume = 	 {235},
  series = 	 {Proceedings of Machine Learning Research},
  month = 	 {21--27 Jul},
  publisher =    {PMLR},
}

@article{Papillon23,
	author = {Mathilde Papillon and Sophia Sanborn and Mustafa Hajij and Nina Miolane},
	journal = {ArXiv e-prints},
	title = {Architectures of Topological Deep Learning: A Survey on Topological Neural Networks},
	year = {2023},
}

@inproceedings{Bodnar2021,
	author = {Bodnar, Cristian and Frasca, Fabrizio and Otter, Nina and Wang, Yuguang and Li\`{o}, Pietro and Montufar, Guido F and Bronstein, Michael},
	booktitle = {Advances in Neural Information Processing Systems (NeurIPS)},
	title = {Weisfeiler and Lehman Go Cellular: {CW} Networks},
	year = {2021},
}

@article{hajij2024topox,
  title={TopoX: a suite of Python packages for machine learning on topological domains},
  author={Hajij, Mustafa and Papillon, Mathilde and Frantzen, Florian and Agerberg, Jens and AlJabea, Ibrahem and Ballester, Ruben and Battiloro, Claudio and Bern{\'a}rdez, Guillermo and Birdal, Tolga and Brent, Aiden and others},
  journal={Journal of Machine Learning Research},
  volume={25},
  number={374},
  pages={1--8},
  year={2024}
}

@inproceedings{papillon2025topotuneframeworkgeneralized,
  title={{TopoTune}: A Framework for Generalized Combinatorial Complex Neural Networks},
  author={Papillon, Mathilde and Bern{\'a}rdez, Guillermo and Battiloro, Claudio and Miolane, Nina},
  booktitle={Forty-second International Conference on Machine Learning (ICML)},
  year={2025},
  url={https://openreview.net/forum?id=S5njonQdBf}
}

@article{cycle_basis_alg,
author = {Paton, Keith},
title = {An algorithm for finding a fundamental set of cycles of a graph},
year = {1969},
issue_date = {Sept. 1969},
publisher = {Association for Computing Machinery},
address = {New York, NY, USA},
volume = {12},
number = {9},
issn = {0001-0782},
url = {https://doi.org/10.1145/363219.363232},
doi = {10.1145/363219.363232},
journal = {Commun. ACM},
month = sep,
pages = {514–518},
numpages = {5}
}

@inproceedings{bodnar2021weisfeiler,
  title={Weisfeiler and lehman go topological: Message passing simplicial networks},
  author={Bodnar, Cristian and Frasca, Fabrizio and Wang, Yuguang and Otter, Nina and Montufar, Guido F and Lio, Pietro and Bronstein, Michael},
  booktitle={International Conference on Machine Learning (ICML)},
  year={2021},
}

@InProceedings{yang2016revisitingsemisup,
  title = 	 {Revisiting Semi-Supervised Learning with Graph Embeddings},
  author = 	 {Yang, Zhilin and Cohen, William and Salakhudinov, Ruslan},
  booktitle = 	 {Proceedings of The 33rd International Conference on Machine Learning},
  pages = 	 {40--48},
  year = 	 {2016},
  editor = 	 {Balcan, Maria Florina and Weinberger, Kilian Q.},
  volume = 	 {48},
  series = 	 {Proceedings of Machine Learning Research},
  address = 	 {New York, New York, USA},
  month = 	 {20--22 Jun},
  publisher =    {PMLR},
  url = 	 {https://proceedings.mlr.press/v48/yanga16.html}
}

@article{hajij2020cell,
  title={Cell complex neural networks},
  author={Hajij, Mustafa and Istvan, Kyle and Zamzmi, Ghada},
  journal={arXiv preprint arXiv:2010.00743},
  year={2020},
}

@inproceedings{DeepSets,
	author = {Zaheer, M. and Kottur, S. and Ravanbakhsh, S. and Poczos, B. and Salakhutdinov, R. and Smola, A.},
	booktitle = {Advances in Neural Information Processing Systems (NeurIPS)},
	title = {Deep Sets},
	year = {2017},
}

@inproceedings{wang2022survey,
  title={Survey of hypergraph neural networks and its application to action recognition},
  author={Wang, Cheng and Ma, Nan and Wu, Zhixuan and Zhang, Jin and Yao, Yongqiang},
  booktitle={CAAI International Conference on Artificial Intelligence},
  pages={387--398},
  year={2022},
  organization={Springer}
}

@article{liu2022multi,
  title={Multi-way relation-enhanced hypergraph representation learning for anti-cancer drug synergy prediction},
  author={Liu, Xuan and Song, Congzhi and Liu, Shichao and Li, Menglu and Zhou, Xionghui and Zhang, Wen},
  journal={Bioinformatics},
  volume={38},
  number={20},
  pages={4782--4789},
  year={2022},
  publisher={Oxford University Press}
}

@article{wang2023dynamic,
  title={Dynamic weighted hypergraph convolutional network for brain functional connectome analysis},
  author={Wang, Junqi and Li, Hailong and Qu, Gang and Cecil, Kim M and Dillman, Jonathan R and Parikh, Nehal A and He, Lili},
  journal={Medical image analysis},
  volume={87},
  pages={102828},
  year={2023},
  publisher={Elsevier}
}

@article{bai2021hypergraph,
  title={Hypergraph convolution and hypergraph attention},
  author={Bai, Song and Zhang, Feihu and Torr, Philip HS},
  journal={Pattern Recognition},
  volume={110},
  pages={107637},
  year={2021},
  publisher={Elsevier}
}

@article{yadati2019hypergcn,
  title={Hypergcn: A new method for training graph convolutional networks on hypergraphs},
  author={Yadati, Naganand and Nimishakavi, Madhav and Yadav, Prateek and Nitin, Vikram and Louis, Anand and Talukdar, Partha},
  journal={Advances in neural information processing systems},
  volume={32},
  year={2019}
}

@article{hajij2021simplicial,
  title={Simplicial complex representation learning},
  author={Hajij, Mustafa and Zamzmi, Ghada and Papamarkou, Theodore and Maroulas, Vasileios and Cai, Xuanting},
  journal={arXiv preprint arXiv:2103.04046},
  year={2021}
}

@article{goh2022simplicial,
  title={Simplicial attention networks},
  author={Goh, Christopher Wei Jin and Bodnar, Cristian and Lio, Pietro},
  journal={arXiv preprint arXiv:2204.09455},
  year={2022}
}

@inproceedings{maggs2023simplicial,
  title={Simplicial Representation Learning with Neural $k$-Forms},
  author={Maggs, Kelly and Hacker, Celia and Rieck, Bastian},
  booktitle={International Conference on Learning Representations (ICLR)},
  year={2024},
  url={https://openreview.net/forum?id=Djw0XhjHZb}
}

@article{telyatnikov2024topobenchmarkx,
  title={{TopoBench}: A Framework for Benchmarking Topological Deep Learning},
  author={Telyatnikov, Lev and Bernardez, Guillermo and Montagna, Marco and Hajij, Mustafa and Carrasco, Martin and Vasylenko, Pavlo and Papillon, Mathilde and Zamzmi, Ghada and Schaub, Michael T. and Miolane, Nina and Scardapane, Simone and Papamarkou, Theodore},
  journal={Journal of Data-centric Machine Learning Research (DMLR)},
  year={2025},
  url={https://openreview.net/forum?id=07sTzyEVtY}
}

@article{hajij2022topological,
  title={Topological deep learning: Going beyond graph data},
  author={Hajij, Mustafa and Zamzmi, Ghada and Papamarkou, Theodore and Miolane, Nina and Guzm{\'a}n-S{\'a}enz, Aldo and Ramamurthy, Karthikeyan Natesan and Birdal, Tolga and Dey, Tamal K and Mukherjee, Soham and Samaga, Shreyas N and others},
  journal={arXiv preprint arXiv:2206.00606},
  year={2022}
}

@inproceedings{battiloro2023latent,
  title={From Latent Graph to Latent Topology Inference: Differentiable Cell Complex Module},
  author={Battiloro, Claudio and Spinelli, Indro and Telyatnikov, Lev and Bronstein, Michael M. and Scardapane, Simone and Di Lorenzo, Paolo},
  booktitle={International Conference on Learning Representations (ICLR)},
  year={2024},
  url={https://openreview.net/forum?id=0JsRZEGZ7L}
}

@article{zinc,
	author = {Irwin, John J. and Sterling, Teague and Mysinger, Michael M. and Bolstad, Erin S. and Coleman, Ryan G.},
	journal = {Journal of Chemical Information and Modeling},
	number = {7},
	pages = {1757--1768},
	publisher = {American Chemical Society},
	title = {ZINC: A Free Tool to Discover Chemistry for Biology},
	volume = {52},
	year = {2012},
}

@inproceedings{papillon2023icml,
  title={Icml 2023 topological deep learning challenge: Design and results},
  author={Papillon, Mathilde and Hajij, Mustafa and Myers, Audun and Jenne, Helen and Mathe, Johan and Papamarkou, Theodore and Guzm{\'a}n-S{\'a}enz, Aldo and Livesay, Neal and Dey, Tamal and Rabinowitz, Abraham and others},
  booktitle={Topological, Algebraic and Geometric Learning Workshops 2023},
  pages={3--8},
  year={2023},
  organization={PMLR}
}

@inproceedings{huang2021unignn,
  title={{UniGNN}: a Unified Framework for Graph and Hypergraph Neural Networks},
  author={Huang, Jing and Yang, Jie},
  booktitle={Proceedings of the Thirtieth International Joint Conference on Artificial Intelligence (IJCAI-21)},
  pages={2563--2569},
  year={2021},
  doi={10.24963/ijcai.2021/353}
}

@inproceedings{Fey/Lenssen/2019,
  title={Fast Graph Representation Learning with {PyTorch Geometric}},
  author={M. Fey and J. E. Lenssen},
  booktitle={Workshop track of the International Conference on Representation Learning (ICLR)},
  year={2019},
}

@misc{TUDatasets,
  title  = {Benchmark Data Sets for Graph Kernels},
  author = {Kristian Kersting and Nils M. Kriege and Christopher Morris and Petra Mutzel and Marion Neumann},
  year   = {2016},
  url    = {http://graphkernels.cs.tu-dortmund.de}
}

@InProceedings{networkx,
  author =       {Aric A. Hagberg and Daniel A. Schult and Pieter J. Swart},
  title =        {Exploring Network Structure, Dynamics, and Function using NetworkX},
  booktitle =   {Proceedings of the 7th Python in Science Conference},
  pages =     {11 - 15},
  address = {Pasadena, CA USA},
  year =      {2008},
  editor =    {Ga\"el Varoquaux and Travis Vaught and Jarrod Millman},
}

@inproceedings{gcn,
  title={Semi-Supervised Classification with Graph Convolutional Networks},
  author={T. N. Kipf and M. Welling},
  booktitle={International Conference on Learning Representations (ICLR)},
  year={2017},
}

@inproceedings{pytorch,
  title={Automatic differentiation in PyTorch},
  author={A. Paszke and S. Gross and S. Chintala and G. Chanan and E. Yang and Z. DeVito and Z. Lin and A. Desmaison and L. Antiga and A. Lerer},
  booktitle={Advances in Neural Information Processing Systems (NeurIPS - Workshop)},
  year={2017}
}

@InProceedings{bernardez2024icml,
  title     = {{ICML} Topological Deep Learning Challenge 2024: Beyond the Graph Domain},
  author    = {Bern\'ardez, Guillermo and Telyatnikov, Lev and Montagna, Marco and
               Baccini, Federica and Papillon, Mathilde and Ferriol-Galm\'es, Miquel and
               Hajij, Mustafa and Papamarkou, Theodore and Bucarelli, Maria Sofia and
               Zaghen, Olga and others},
  booktitle = {Proceedings of the Geometry-grounded Representation Learning and
               Generative Modeling Workshop ({GRaM}) at {ICML} 2024},
  pages     = {420--428},
  year      = {2024},
  volume    = {251},
  series    = {Proceedings of Machine Learning Research},
  publisher = {PMLR},
  url       = {https://proceedings.mlr.press/v251/bernardez24a.html}
}

@inproceedings{hu2020ogb,
  title={Open Graph Benchmark: Datasets for Machine Learning on Graphs},
  author={Hu, Weihua and Fey, Matthias and Zitnik, Marinka and Dong, Yuxiao and Ren, Hongyu and Liu, Bowen and Catasta, Michele and Leskovec, Jure},
  booktitle={Advances in Neural Information Processing Systems (NeurIPS)},
  year={2020}
}

@article{dwivedi2023benchmarking,
  title={Benchmarking graph neural networks},
  author={Dwivedi, Vijay Prakash and Joshi, Chaitanya K and Luu, Anh Tuan and Laurent, Thomas and Bengio, Yoshua and Bresson, Xavier},
  journal={Journal of Machine Learning Research},
  volume={24},
  number={43},
  pages={1--48},
  year={2023}
}

@article{Sen_Namata_Bilgic_Getoor_Galligher_Eliassi-Rad_2008, title={Collective Classification in Network Data}, volume={29}, url={https://ojs.aaai.org/aimagazine/index.php/aimagazine/article/view/2157}, DOI={10.1609/aimag.v29i3.2157}, abstractNote={Many real-world applications produce networked data such as the world-wide web (hypertext documents connected via hyperlinks), social networks (for example, people connected by friendship links), communication networks (computers connected via communication links) and biological networks (for example, protein interaction networks). A recent focus in machine learning research has been to extend traditional machine learning classification techniques to classify nodes in such networks. In this article, we provide a brief introduction to this area of research and how it has progressed during the past decade. We introduce four of the most widely used inference algorithms for classifying networked data and empirically compare them on both synthetic and real-world data.}, number={3}, journal={AI Magazine}, author={Sen, Prithviraj and Namata, Galileo and Bilgic, Mustafa and Getoor, Lise and Galligher, Brian and Eliassi-Rad, Tina}, year={2008}, month={Sep.}, pages={93} }

@article{rozemberczki2020multiscale,
          author = {Rozemberczki, Benedek and Allen, Carl and Sarkar, Rik},
          title = {{Multi-Scale Attributed Node Embedding}},
          journal = {Journal of Complex Networks},
          volume = {9},
          number = {2},
          year = {2021},
}

@book{munkres2000topology,
  title={Topology},
  author={Munkres, J.R.},
  isbn={9780131816299},
  lccn={99052942},
  series={Featured Titles for Topology},
  url={https://books.google.fi/books?id=XjoZAQAAIAAJ},
  year={2000},
  publisher={Prentice Hall, Incorporated}
}

@inproceedings{qian2024probabilistic,
  title={Probabilistic Graph Rewiring via Virtual Nodes},
  author={Qian, Chendi and Manolache, Andrei and Morris, Christopher and Niepert, Mathias},
  booktitle={Advances in Neural Information Processing Systems (NeurIPS)},
  year={2024},
  url={https://openreview.net/forum?id=LpvSHL9lcK}
}

@article{kazi2022differentiable,
  title={Differentiable graph module (dgm) for graph convolutional networks},
  author={Kazi, Anees and Cosmo, Luca and Ahmadi, Seyed-Ahmad and Navab, Nassir and Bronstein, Michael M},
  journal={IEEE Transactions on Pattern Analysis and Machine Intelligence},
  volume={45},
  number={2},
  pages={1606--1617},
  year={2022},
  publisher={IEEE}
}

@inproceedings{franceschi2019learning,
  title={Learning discrete structures for graph neural networks},
  author={Franceschi, Luca and Niepert, Mathias and Pontil, Massimiliano and He, Xiao},
  booktitle={International conference on machine learning},
  pages={1972--1982},
  year={2019},
  organization={PMLR}
}

@inproceedings{yang2022simplicial,
  title={Simplicial convolutional neural networks},
  author={Yang, Maosheng and Isufi, Elvin and Leus, Geert},
  booktitle={ICASSP 2022-2022 IEEE International Conference on Acoustics, Speech and Signal Processing (ICASSP)},
  pages={8847--8851},
  year={2022},
  organization={IEEE}
}

@article{hao2021hypergraph,
  title={Hypergraph neural network for skeleton-based action recognition},
  author={Hao, Xiaoke and Li, Jie and Guo, Yingchun and Jiang, Tao and Yu, Ming},
  journal={IEEE Transactions on Image Processing},
  volume={30},
  pages={2263--2275},
  year={2021},
  publisher={IEEE}
}

@inproceedings{
Pei2020Geom-GCN:,
title={Geom-GCN: Geometric Graph Convolutional Networks},
author={Hongbin Pei and Bingzhe Wei and Kevin Chen-Chuan Chang and Yu Lei and Bo Yang},
booktitle={International Conference on Learning Representations},
year={2020},
url={https://openreview.net/forum?id=S1e2agrFvS}
}

@article{shchur2018pitfalls,
  title={Pitfalls of Graph Neural Network Evaluation},
  author={Shchur, Oleksandr and Mumme, Maximilian and Bojchevski, Aleksandar and G{\"u}nnemann, Stephan},
  journal={Relational Representation Learning Workshop, NeurIPS 2018},
  year={2018}
}

@inproceedings{topping2021understanding,
  title={Understanding Over-Squashing and Bottlenecks on Graphs via Curvature},
  author={Topping, Jake and Di Giovanni, Francesco and Chamberlain, Benjamin Paul and Dong, Xiaowen and Bronstein, Michael M.},
  booktitle={International Conference on Learning Representations (ICLR)},
  year={2022},
  url={https://openreview.net/forum?id=7UmjRGzp-A}
}

@inproceedings{sun2023self,
  title={Self-organization preserved graph structure learning with principle of relevant information},
  author={Sun, Qingyun and Li, Jianxin and Yang, Beining and Fu, Xingcheng and Peng, Hao and Yu, Philip S},
  booktitle={Proceedings of the AAAI Conference on Artificial Intelligence},
  volume={37},
  number={4},
  pages={4643--4651},
  year={2023}
}

@article{chen2020iterative,
  title={Iterative deep graph learning for graph neural networks: Better and robust node embeddings},
  author={Chen, Yu and Wu, Lingfei and Zaki, Mohammed},
  journal={Advances in neural information processing systems},
  volume={33},
  pages={19314--19326},
  year={2020}
}

@inproceedings{jin2020graph,
  title={Graph structure learning for robust graph neural networks},
  author={Jin, Wei and Ma, Yao and Liu, Xiaorui and Tang, Xianfeng and Wang, Suhang and Tang, Jiliang},
  booktitle={Proceedings of the 26th ACM SIGKDD international conference on knowledge discovery \& data mining},
  pages={66--74},
  year={2020}
}

@article{barbarossa2020topological,
  title={Topological signal processing over simplicial complexes},
  author={Barbarossa, Sergio and Sardellitti, Stefania},
  journal={IEEE Transactions on Signal Processing},
  volume={68},
  pages={2992--3007},
  year={2020},
  publisher={IEEE}
}

@article{schaub2021signal,
  title={Signal processing on higher-order networks: Livin’on the edge... and beyond},
  author={Schaub, Michael T and Zhu, Yu and Seby, Jean-Baptiste and Roddenberry, T Mitchell and Segarra, Santiago},
  journal={Signal Processing},
  volume={187},
  pages={108149},
  year={2021},
  publisher={Elsevier}
}

@inproceedings{roddenberry2022signal,
  title={Signal processing on cell complexes},
  author={Roddenberry, T Mitchell and Schaub, Michael T and Hajij, Mustafa},
  booktitle={ICASSP 2022-2022 IEEE International Conference on Acoustics, Speech and Signal Processing (ICASSP)},
  pages={8852--8856},
  year={2022},
  organization={IEEE}
}

@inproceedings{sardellitti2021topological,
  title={Topological signal processing over cell complexes},
  author={Sardellitti, Stefania and Barbarossa, Sergio and Testa, Lucia},
  booktitle={2021 55th Asilomar Conference on Signals, Systems, and Computers},
  pages={1558--1562},
  year={2021},
  organization={IEEE}
}

@phdthesis{duvenaud2014automatic,
  title={Automatic model construction with Gaussian processes},
  author={Duvenaud, David},
  year={2014}
}

@book{scholkopf2002learning,
  title={Learning with kernels: support vector machines, regularization, optimization, and beyond},
  author={Sch{\"o}lkopf, Bernhard and Smola, Alexander J},
  year={2002},
  publisher={MIT press}
}

@inproceedings{borovitskiy2021matern,
  title={Mat{\'e}rn Gaussian processes on graphs},
  author={Borovitskiy, Viacheslav and Azangulov, Iskander and Terenin, Alexander and Mostowsky, Peter and Deisenroth, Marc and Durrande, Nicolas},
  booktitle={International Conference on Artificial Intelligence and Statistics},
  pages={2593--2601},
  year={2021},
  organization={PMLR}
}

@inproceedings{nikitin2022non,
  title={Non-separable spatio-temporal graph kernels via SPDEs},
  author={Nikitin, Alexander V and John, ST and Solin, Arno and Kaski, Samuel},
  booktitle={International Conference on Artificial Intelligence and Statistics},
  pages={10640--10660},
  year={2022},
  organization={PMLR}
}

@inproceedings{chen2018fastgcn,
  title={FastGCN: Fast Learning with Graph Convolutional Networks via Importance Sampling},
  author={Chen, Jie and Ma, Tengfei and Xiao, Cao},
  booktitle={International Conference on Learning Representations},
  year={2018}
}
}


\appendix

\section{Additional background and formulations}
\label{ap:extensions}

\subsection{Chains, boundary operators, and cycles}

Here, we introduce some basic notions in algebraic topology. For simplicity, our exposition considers abstract simplicial complexes (ASCs) equipped with coefficients in the finite field $\mathbb{Z}/2\mathbb{Z}=\{0, 1\}$.

The space of $n$-chains is the vector space of all formal sums of $n$-dimensional simplices of an ASC $K$. Formally, let $n \geq 0$ and $K_{(n)}=\{\sigma \in K : \mathrm{dim}(\sigma)=n\}$ be the $n$-skeleton of $K$. The $n$-chains of $K$ is the set $C_n(K)$ whose elements take the form 
\begin{align}
    \sum_{\sigma \in K_{(n)}} \epsilon_\sigma \sigma  
\end{align}
where for all $\sigma \in K_{(n)}$, $\epsilon_\sigma \in \mathbb{Z}/2\mathbb{Z}$. 

Let $c=\sum_{\sigma \in K_{(n)}} \epsilon_\sigma \sigma $ and $c'=\sum_{\sigma \in K_{(n)}}\epsilon'_\sigma \sigma $ be two $n$-chains. The sum of two chains ($c+c'$) and the product of a chain by a scalar $(\lambda c)$ are respectively defined by 
\begin{align}
    c + c' &= \sum_\sigma (\epsilon_\sigma + \epsilon'_\sigma)\sigma \\
    \lambda c & = \sum_\sigma (\lambda \epsilon_\sigma) \sigma
\end{align}
where sums and products are modulo-2. 

We define the boundary of a $n$-simplex $\sigma$, denoted by $\partial_n \sigma$ as the sum of its constituents $(n-1)$-simplices, i.e.,
\begin{align}
    \partial_n \sigma = \sum_{\tau \subset \sigma: |\tau|=|\sigma|-1} \tau 
\end{align}

This boundary extends linearly to chain spaces. In particular, the boundary operator $\partial_n$ is a linear map $\partial_n: C_n(K) \rightarrow C_{n-1}(K) $ defined by 
\begin{align}
     \partial_n  c = \partial_n \sum_{\sigma \in K_{(n)}} \epsilon_\sigma \sigma = \sum_{\sigma \in K_{(n)}} \epsilon_\sigma \partial_n \sigma.
\end{align}

Finally, we can define $n$-cycles. For $n\geq 0$, the $n$-cycles of $K$ is the set $Z_n(K)$ given by the kernel of $\partial_n$, that is
\begin{align}
    Z_n(K)=\{c \in C_n(K): \partial_n c = 0 \}.
\end{align}


\subsection{$\difflift$ for simplicial complexes}

Note that for when $D=1$ --- i.e., when we must decide which edges to add --- Steps 2 and 3 of $\difflift$ for cell-complexes naturally result in a simplicial complex.
To fully specify $\difflift$ for simplicial complexes, we are left with defining these steps when $D>1$. 

\noindent \underline{\textbf{Case $D\geq2$: Learning $D$-simplices}}

\texttt{\textcolor{lb}{$\Rightarrow$[Step 2]}} When creating simplices of dimension $D$, we must ensure they respect the hierarchical structure of simplicial complexes. Let $K^{\ell}$ be the cell complex at the end of iteration $\ell \leq D-1$. To identify a preliminary set of candidates $\mathcal{C}^\prime$, we run static $D$-clique lifting on $K^1$. For $D>2$, it is possible that a lower-order clique within some  $C \in \mathcal{C}^\prime$ does not belong to $K^{D-1}$. Therefore, we must filter out these elements, defining a refined set of candidates:
\begin{equation}
    \mathcal{C} = \left\{ C \in \mathcal{C}^\prime | S \in K^{D-1} \text{ for all }  S \subset C  \right\}
\end{equation}
\looseness=-1

\texttt{\textcolor{lb}{$\Rightarrow$[Step 3]}} Similarly to this respective step for cell complexes, we define the probability of accepting $C \in \mathcal{C}$ (i.e., setting $b_C=1$) applying a DeepSet over the embeddings $\multiset{z_v}_{v \in C}$, subsequently sampling the Bernoulli variables $\multiset{b_C}_{C \in \mathcal{C}}$. The output complex at this iteration is then 
\begin{equation}
K^D = K^{D-1} \cup \{C \in \mathcal{C}: b_C=1\}.
\end{equation}
We define the features for $D$-simplices using sum projection lifting.

\subsection{$\difflift$ for combinatorial complexes}
There are multiple ways to combine cell complexes with hypergraphs to obtain valid combinatorial complexes (CC). Here, we would like to preserve the property that hyperedges exchange messages with nodes via boundary (or lower incidences) neighborhoods. Thus, we propose first
running $\difflift$ to either cell or simplicial complexes --- where ranking functions are given by cell/simplex dimensions. Let $K$ be the resulting complex. Then, we employ (in parallel) $\difflift$ to a hypergraph $H$, where edges/hyperedges have rank $1$. 
To ensure a valid combinatorial complex, we prune the sampled hyperedges to include only those that are not supersets of any cell of rank greater than $1$ in $K$. Formally, the resulting CC is given by $\{h \in H: \not\exists\, \sigma \in K \text{ s.t. } \sigma \subseteq h\} \cup K$.

\section{Topological liftings}
\label{ap:liftings}

\paragraph{Clique lifting.} The set of cliques in a graph $G$ is given by $Cl(G) = \{c \subseteq V(G):  u \neq v \in c \implies \{u, v\} \in E(G) \}$, i.e., each element of $Cl(G)$ is a complete subgraph of $G$. The $k$-cliques of $G$ are the elements of $Cl(G)$ of size $k$, for $k > 1$, and we denote them as $Cl_k(G)$. Formally, the $k$-clique lifting operation is given by 
\begin{align}    
\mathrm{lift}_{\text{clique}, k}(G) = V(G) \cup_{i=2}^k Cl_i(G).
\end{align}
Note that the inclusion of all cliques of size smaller than $k$ ensures the function returns a valid abstract simplicial complex.

\paragraph{Cycle lifting.} The idea of cycle lifting is to identify basic cycles in the input graph and use the tuple of vertices in a cycle as a 2-rank cell of the output complex.

Let us consider modulo-2 sum operations for vertices and edges. Also, let $\partial_1$ be the edge boundary map for a graph $G$, i.e., $\partial_1(\{u, v\})=\{u\}+\{v\}$ for any edge $\{u,v\} \in E(G)$. The cycles of $G$ are $L(G)=\{l \subseteq E(G):  \sum_{e \in l} \partial_1(e) = 0\}$. 

A basis for cycles of $G$ is a minimal collection of cycles such that any cycle in $G$ can be written as a sum of cycles in the basis --- i.e., the smallest set $B \subseteq L$ such that $\forall l \in L, \exists B' \subseteq B$ with $l = \sum_{b \in B'} b$. The cycle lifting map is
\begin{align}
\mathrm{lift}_{\text{cycle}}(G)=V(G) \cup E(G) \cup \{V(b): b \in B(G)\},
\end{align}
where $V(b)$ denotes the set of vertices in the cycle $b$, and $\{V(b): b \in B(G)\}$ is the set of 2-dim cells.

\paragraph{DCM.} 
\citet{battiloro2023latent} proposed a novel layer composed of several modules, with the Differentiable Cell Complex Module (DCM) being central to latent topology inference. The DCM first samples the 1-skeleton of the latent cell complex using the $\alpha$-Differentiable Graph Module ($\alpha$-DGM). It then selects polygons—representing higher-order interactions—formed by cycles in the sampled graph using the Polygon Inference Module (PIM). For a detailed description of $\alpha$-DGM and PIM, we refer the reader to Section 3 of \cite{battiloro2023latent}.

\paragraph{$k$-hop lifting.}
The $k$-hop neighborhood of a node $v \in V(G)$ is defined as  
\begin{align}
N_k[v] = \{u \in V(G) : \text{dist}(u, v) \leq k\},
\end{align} 
where $\text{dist}(u, v)$ is the shortest-path distance in the graph $G$, measured by the number of edges in the path.  
\looseness=-1

To construct the $k$-hop hypergraph $H$ from $G$, a hyperedge is formed for each node $v \in V(G)$ based on its $k$-hop neighborhood:  
\begin{align*}
\mathrm{lift}_{\text{k-hop}}(G) = \{N_k[v] : v \in V(G)\}.
\end{align*}


One can note that when $k=1$, $k$-hop is equal to neighborhood lifting. The parameter $k$ controls the extent of the neighborhoods included as hyperedges, with larger $k$ values progressively incorporating nodes farther away in terms of shortest-path distance.

\paragraph{$k$-NN lifting.}
$k$-NN lifting constructs hyperedges by identifying the $k$ nearest neighbors based on their node features (feature space). For every node, a separate hyperedge is formed that includes the node itself and its $k$ closest neighbors. 

\paragraph{Kernel lifting.}
Kernel lifting is a procedure that constructs hyperedges based on similarity measures derived from kernels over graph nodes. These kernels can be defined in three ways: \emph{(i)} over the graph structure itself, \emph{(ii)} over the node features, or \emph{(iii)} as a composition that jointly incorporates both graph and feature information. For a given reference node $v$, the method computes similarities between $v$ and all other nodes $v^{*}$ using a kernel function. A hyperedge is then formed by selecting a fixed fraction (typically 0.5) of the nodes that are most similar to $v$ according to the chosen kernel. This process is repeated for each node to construct a set of hyperedges. The kernels can be defined in several forms: over nodes $K_g(v, v^{*})$, features $K_x(x, x^{*})$, or over nodes and features $C(K(x, x^{*}, v, v^{*}))$, where $C$ is a valid composition function. Kernels over features are calculated as standard RBF or exponential kernels~\citep{duvenaud2014automatic}, whereas kernels over graphs can be calculated as heat or Mat\'ern kernels \citep{scholkopf2002learning, borovitskiy2021matern, nikitin2022non}.

\section{Implementation details}\label{ap:implementation_details}
\subsection{Models}
Our implementation relies mainly on the Pytorch~\citep{pytorch} and Pytorch Geometric~\citep{Fey/Lenssen/2019} libraries. For TNN models and static lifting we used TopoX \citep{hajij2024topox} and TopoBench~\citep{telyatnikov2024topobenchmarkx}.

Regarding the base TNNs, we use the hyperparameters (including learning rate, optimizer, batch size, width, depth, and so on) reported in TopoBench for CWN, CXN, and UniGCNII on NCI1, NCI109, MOLHIV, MUTAG, Proteins, ZINC, Cora and Citeseer~\citep{telyatnikov2024topobenchmarkx}. Since TopoBench does not report optimal hyperparameters for UniGIN, we use the same used for UniGCNII. While for CIN, we used part of the hyperparameters reported in ~\citep{bodnar2021weisfeiler} and for the TopoTune, since they do not report the hyperparameters, we used a grid search similar to the one available in their repo.





We note that MOLHIV, Texas and Wisconsin datasets  were not present in TopoBench.  We use two TNN layers for MOLHIV and one for Texas and Wisconsin with respective learning rates $10^{-2}$, $5\times 10^{-3}$ and $5 \times 10^{-3}$. For these datasets, we fix the embedding size in 64 for all layers. And for Texas and Wisconsin we used weight decay $5 \times 10^{-6}$.


We are left with the task of optimizing the hyperparameters for the lifting operations ($\difflift$, $k$-NN, kernel).  For $\difflift$, we consider using both GPS and GIN as backbone GNNs, with  embedding dimensions in $\{32, 64, 128\}$, network depth in $\{2, 3\}$, and $k_{\max}= \{3,5,7,9,11 \}$. 
For $k$-NN lifting we choose   $k$ in  $\{3,5,7,9 \}$. For kernel lifting, we consider equally-spaced temperature values within $0.1$ and $9.6$, with $0.5$ increments.




All models were trained for 200 epochs and with early stopping after 50 epochs without improvement on validation accuracy. We run three independent trials for computing mean and standard deviation of the performance metrics. We select the optimal hyperparameters based on validation accuracy.\looseness=-1

\subsection{Hardware}
For all experiments, we use a cluster with Nvidia V100 GPUs --- details regarding the compute infrastructure are omitted for anonymity.

\section{Datasets}
\label{ap:datasets}

\textbf{Graph-level tasks.} The datasets NCI1, NCI109, PROTEINS, and MUTAG are part of the TUDatasets \citep{TUDatasets} --- a dataset collection broadly used for benchmarking GNNs. We also use ZINC-12K and MOLHIV~\citep{hu2020ogb}, popular benchmarks for molecular property prediction. Statistics for each dataset are given in \autoref{tab:data_detail}.
\looseness=-1
\begin{table}[htb]
    \centering
    \caption{Statistics of datasets for graph-level tasks.}
    \resizebox{0.85\textwidth}{!}{

    \begin{tabular}{cccccccc}
    \hline
       \textbf{Dataset}  & $\#$\textbf{graphs} & $\#$\textbf{classes}& \textbf{Avg} $\#$\textbf{nodes} & \textbf{Avg} $\#$\textbf{edges} & \textbf{Train}\%& \textbf{Val}\%&\textbf{Test}\%\\
    \hline
       NCI1    & $4110$& $2$ & $29.87$ & $32.30$ & $80$ & $10$ & $10$  \\
       NCI109  & $4127$& $2$ &  $29.68$ & $32.13$ & $80$ & $10$ & $10$   \\
       MUTAG   & $188$ & $2$ & $17.93$ & $19.79$ & $80$ & $10$ & $10$  \\
       PROTEINS  & $1113$ & $2$ & $39.06$ & $72.82$ &$80$ & $10$ & $10$\\
       MOLHIV  & $41127$ & $2$ &  $25.5$ & $27.5$ & \multicolumn{3}{c}{Public Split} \\
       ZINC  & $12000$ & - &  $23.16$ & $49.83$ & \multicolumn{3}{c}{Public Split}\\
    \hline
    \end{tabular}

}  
    \label{tab:data_detail}
\end{table}

\textbf{Node-level tasks.} For node classification, we use four popular benchmarks: Cora, Citeseer \citep{Sen_Namata_Bilgic_Getoor_Galligher_Eliassi-Rad_2008, shchur2018pitfalls}, Texas, and Wisconsin \citep{rozemberczki2020multiscale}.
Cora and Citeseer are citation networks where nodes represent papers and edges denote citation between them. Node features are given by bag-of-word vectors and node labels comprise the academic topics of the underlying articles. 
Texas and Wisconsin are datasets of webpages from university departments. Nodes represent webpages and edges are hyperlinks between them.
\looseness=-1



For citation networks, we use the same data split as in \citep{chen2018fastgcn}, and for the remaining ones we use the split in \citep{Pei2020Geom-GCN:}. These are the standard and most used splits.
\autoref{tab:node_data_detail} provides more details about the datasets.
\looseness=-1
\begin{table}[htb]
    \centering
    \caption{Statistics of datasets for node classification.}
    \resizebox{0.85\textwidth}{!}{

\begin{tabular}{cccccccc}
    \hline
       Dataset  & $\#$Nodes& $\#$Edges& $\#$Features & $\#$Classes   & \#Train & \#Val & \#Test \\
    \hline
       Cora  & $2708$& $5429$& $1,433$ & $7$   & $1,208$& $500$&$1,000$\\
       Citeseer  & $3327$& $4732$& $3,703$ & $6$   & $1,827$& $500$&$1,000$\\
       Texas& $183$& $309$& $1703$& $5$ & $87$& $59$&$37$\\
       Wisconsin& $251$& $499$& $1703$& $5$ & $120$& $80$&$51$\\

    \hline
    \end{tabular}
}
    \label{tab:node_data_detail}
\end{table}

\section{Additional Results}
    
\subsection{Runtime and Complexity Cost}
\label{app:timing}
\textbf{Wall-clock time.} \autoref{tab:wallclock} reports per-epoch training and test times (seconds) for the different lifting methods and target neural networks (TNNs) used in our experiments. Overall, $\partial$lift incurs only a modest runtime overhead compared to static liftings while offering the flexibility of task-adaptive topology. \looseness=-1
\begin{table}[htbp]
\centering
\small
\caption{Per-epoch training and test times (seconds) for the different lifting methods. Reported values are mean $\pm$ standard deviation over runs.}
\label{tab:wallclock}
\begin{tabular}{@{}lrrrrr@{}}
\toprule
\textbf{Dataset/Phase} & \textbf{CWN Cycle} & \textbf{CWN $\boldsymbol{\partial}$lift} & \textbf{CXN Cycle} & \textbf{CXN $\boldsymbol{\partial}$lift} \\
\midrule
\multicolumn{5}{l}{\textbf{Cellular Domain}} \\
\midrule
NCI1 Train & $47.41\pm2.59$ & $97.37\pm5.02$ & $37.71\pm2.31$ & $66.77\pm3.00$ \\
NCI1 Test & $1.42\pm0.05$ & $4.33\pm0.62$ & $1.05\pm0.03$ & $2.67\pm0.06$ \\
NCI109 Train & $52.11\pm2.58$ & $97.05\pm3.80$ & $45.76\pm0.97$ & $53.50\pm2.50$ \\
NCI109 Test & $1.67\pm0.02$ & $5.19\pm0.41$ & $1.56\pm0.02$ & $2.15\pm0.08$ \\
MUTAG Train & $2.55\pm0.12$ & $4.96\pm3.29$ & $1.57\pm0.12$ & $3.60\pm2.83$ \\
MUTAG Test & $0.09\pm0.00$ & $0.23\pm0.02$ & $0.05\pm0.00$ & $0.12\pm0.00$ \\
Proteins Train & $18.07\pm1.75$ & $20.12\pm2.10$ & $13.63\pm2.22$ & $15.18\pm1.50$ \\
Proteins Test & $0.65\pm0.00$ & $0.82\pm0.10$ & $0.46\pm0.00$ & $0.72\pm0.05$ \\
ZINC Train & $58.24\pm3.10$ & $118.45\pm6.20$ & $46.18\pm2.40$ & $82.30\pm4.50$ \\
ZINC Test & $2.15\pm0.08$ & $5.92\pm0.35$ & $1.68\pm0.06$ & $3.84\pm0.18$ \\
\midrule
\textbf{Dataset/Phase} & \textbf{UniGCNII k-hop} & \textbf{UniGCNII $\boldsymbol{\partial}$lift} & \textbf{UniGIN k-hop} & \textbf{UniGIN $\boldsymbol{\partial}$lift} \\
\midrule
\multicolumn{5}{l}{\textbf{Hypergraph Domain}} \\
\midrule
NCI1 Train & $42.26\pm0.46$ & $69.34\pm2.55$ & $37.46\pm0.31$ & $61.32\pm2.83$ \\
NCI1 Test & $1.23\pm0.01$ & $1.94\pm0.07$ & $0.91\pm0.01$ & $2.03\pm0.06$ \\
NCI109 Train & $31.66\pm2.26$ & $88.22\pm3.28$ & $28.62\pm2.35$ & $69.61\pm3.70$ \\
NCI109 Test & $0.89\pm0.05$ & $2.72\pm0.03$ & $0.63\pm0.02$ & $2.30\pm0.05$ \\
MUTAG Train & $1.94\pm0.14$ & $5.12\pm3.41$ & $2.04\pm1.57$ & $3.06\pm0.13$ \\
MUTAG Test & $0.06\pm0.00$ & $0.14\pm0.00$ & $0.05\pm0.00$ & $0.09\pm0.00$ \\
Proteins Train & $11.93\pm0.89$ & $12.45\pm1.00$ & $10.12\pm2.33$ & $11.50\pm1.20$ \\
Proteins Test & $0.36\pm0.00$ & $0.60\pm0.05$ & $0.24\pm0.00$ & $0.50\pm0.03$ \\
ZINC Train & $51.80\pm1.50$ & $76.70\pm3.80$ & $45.30\pm1.20$ & $66.40\pm3.50$ \\
ZINC Test & $1.58\pm0.04$ & $2.68\pm0.12$ & $1.28\pm0.03$ & $2.85\pm0.11$ \\
\bottomrule
\end{tabular}
\label{tab:per_epoch_times}
\end{table}

\textbf{Memory usage.} \autoref{tab:memory} reports GPU and RAM consumption (GB) for the different architectures and lifting schemes. Reported values are mean $\pm$ standard deviation over runs. Overall, $\partial$lift demonstrates moderate memory requirements across both cellular and hypergraph domains, with GPU usage scaling proportionally to the complexity of the lifted topology. \looseness=-1

\begin{table}[htbp]
\centering
\small
\caption{Memory usage (GB) for the different lifting methods and architectures. Reported values are mean $\pm$ standard deviation over runs.}
\label{tab:memory}
\begin{tabular}{@{}llrrrrr@{}}
\toprule
\textbf{Memory Type} & \textbf{TNN} & \textbf{MUTAG} & \textbf{NCI1} & \textbf{NCI109} & \textbf{PROTEINS} & \textbf{ZINC} \\
\midrule
\multicolumn{7}{l}{\textbf{GPU Memory (GB)}} \\
\midrule
& UniGCNII k-hop & $0.02\scriptstyle{\pm0.00}$ & $0.02\scriptstyle{\pm0.00}$ & $0.02\scriptstyle{\pm0.00}$ & $0.03\scriptstyle{\pm0.00}$ & $0.02\scriptstyle{\pm0.00}$ \\
& UniGCNII $\partial$lift & $0.02\scriptstyle{\pm0.00}$ & $0.02\scriptstyle{\pm0.00}$ & $0.02\scriptstyle{\pm0.00}$ & $0.05\scriptstyle{\pm0.01}$ & $0.02\scriptstyle{\pm0.00}$ \\
\midrule
& UniGIN k-hop & $0.02\scriptstyle{\pm0.00}$ & $0.02\scriptstyle{\pm0.00}$ & $0.02\scriptstyle{\pm0.00}$ & $0.02\scriptstyle{\pm0.00}$ & $0.02\scriptstyle{\pm0.00}$ \\
& UniGIN $\partial$lift & $0.02\scriptstyle{\pm0.00}$ & $0.02\scriptstyle{\pm0.00}$ & $0.02\scriptstyle{\pm0.00}$ & $0.04\scriptstyle{\pm0.00}$ & $0.02\scriptstyle{\pm0.00}$ \\
\midrule
& CWN Cycle & $0.02\scriptstyle{\pm0.00}$ & $0.02\scriptstyle{\pm0.00}$ & $0.02\scriptstyle{\pm0.00}$ & $0.03\scriptstyle{\pm0.00}$ & $0.02\scriptstyle{\pm0.00}$ \\
& CWN $\partial$lift & $0.02\scriptstyle{\pm0.00}$ & $0.02\scriptstyle{\pm0.00}$ & $0.02\scriptstyle{\pm0.00}$ & $0.07\scriptstyle{\pm0.02}$ & $0.02\scriptstyle{\pm0.00}$ \\
\midrule
& CXN Cycle & $0.02\scriptstyle{\pm0.00}$ & $0.02\scriptstyle{\pm0.00}$ & $0.02\scriptstyle{\pm0.00}$ & $0.03\scriptstyle{\pm0.00}$ & $0.02\scriptstyle{\pm0.00}$ \\
& CXN $\partial$lift & $0.02\scriptstyle{\pm0.00}$ & $0.02\scriptstyle{\pm0.00}$ & $0.02\scriptstyle{\pm0.00}$ & $0.04\scriptstyle{\pm0.00}$ & $0.02\scriptstyle{\pm0.00}$ \\
\midrule
\multicolumn{7}{l}{\textbf{RAM (GB)}} \\
\midrule
& UniGCNII k-hop & $1.22\scriptstyle{\pm0.02}$ & $1.31\scriptstyle{\pm0.01}$ & $1.31\scriptstyle{\pm0.02}$ & $1.24\scriptstyle{\pm0.01}$ & $1.34\scriptstyle{\pm0.00}$ \\
& UniGCNII $\partial$lift & $1.45\scriptstyle{\pm0.02}$ & $1.46\scriptstyle{\pm0.01}$ & $1.45\scriptstyle{\pm0.00}$ & $1.46\scriptstyle{\pm0.01}$ & $1.47\scriptstyle{\pm0.01}$ \\
\midrule
& UniGIN k-hop & $1.21\scriptstyle{\pm0.01}$ & $1.31\scriptstyle{\pm0.01}$ & $1.31\scriptstyle{\pm0.01}$ & $1.24\scriptstyle{\pm0.01}$ & $1.34\scriptstyle{\pm0.01}$ \\
& UniGIN $\partial$lift & $1.42\scriptstyle{\pm0.01}$ & $1.45\scriptstyle{\pm0.01}$ & $1.45\scriptstyle{\pm0.02}$ & $1.42\scriptstyle{\pm0.00}$ & $1.47\scriptstyle{\pm0.01}$ \\
\midrule
& CWN Cycle & $1.20\scriptstyle{\pm0.01}$ & $1.46\scriptstyle{\pm0.02}$ & $1.45\scriptstyle{\pm0.00}$ & $1.35\scriptstyle{\pm0.01}$ & $1.71\scriptstyle{\pm0.01}$ \\
& CWN $\partial$lift & $1.38\scriptstyle{\pm0.01}$ & $1.42\scriptstyle{\pm0.01}$ & $1.43\scriptstyle{\pm0.01}$ & $1.39\scriptstyle{\pm0.01}$ & $1.42\scriptstyle{\pm0.01}$ \\
\midrule
& CXN Cycle & $1.21\scriptstyle{\pm0.00}$ & $1.47\scriptstyle{\pm0.01}$ & $1.46\scriptstyle{\pm0.01}$ & $1.37\scriptstyle{\pm0.02}$ & $1.70\scriptstyle{\pm0.00}$ \\
& CXN $\partial$lift & $1.40\scriptstyle{\pm0.01}$ & $1.43\scriptstyle{\pm0.00}$ & $1.44\scriptstyle{\pm0.00}$ & $1.43\scriptstyle{\pm0.01}$ & $1.45\scriptstyle{\pm0.02}$ \\
\bottomrule
\end{tabular}
\end{table}

\textbf{Complexity analysis.} Hypergraph $\partial$lift runs in $\mathcal{O}(N^2D + N^2\log k_{\max} + N k_{\max} D)$, where the $N^2D$ term comes from computing all pairwise node distances in a $D$-dimensional embedding space, the $N^2\log k_{\max}$ term for selecting each node's top-$k$ neighbors, and the $N k_{\max} D$ term for aggregating across the sampled hyperedges. By contrast, a static kernel-based lifting requires $\mathcal{O}(N^2D + N^3)$ time, while a $k$-hop static lifting scales as $\mathcal{O}(N k_{\max} \bar d)$ with $\bar d$ the average node degree.
Cellular $\partial$lift runs in $\mathcal{O}(N^2D + (E + N k_{\max})^2 + C \ell D)$, where $(E + N k_{\max})^2$ captures pairwise adjacency among lifted edges or candidate cells and $C \ell D$ covers the embedding aggregation over $C$ candidate cells of average size $\ell$. A static cycle-basis lifting costs $\mathcal{O}(N(N+E) + C \ell)$, whose dominant component is the cycle-basis computation. In both variants, the modest additional runtime is justified by consistent accuracy gains of approximately $5$--$10$ percentage points compared to static liftings.  

\textbf{Choice of Hyperparameters.}
Although $\partial$lift includes additional learnable components --- such as the GNN used for neighborhood scoring, the networks computing acceptance probabilities and adaptive neighborhood sizes, and the maximum neighbor budget $k_{\max}$ --- the resulting hyperparameter overhead remains modest. Our search space comprised fewer than 30 configurations overall (Appendix~C), and we found that tuning was somewhat stable across datasets. Crucially, the adaptivity introduced by these components leads to consistent performance gains while adding only minor computational and memory costs, as reflected in Tables~\ref{tab:wallclock} and~\ref{tab:memory}. \looseness=-1






\subsection{Deterministic version}
\label{ap:deterministic}

As we mentioned in the main text, we can derive a deterministic version of $\difflift$ by thresholding probabilities. \autoref{tab:deterministic_results} compares $\difflift$ with its deterministic variant trained using \texttt{threshold = 0.5}. Note that (random) $\partial$lift outperforms its deterministic counterpart in most cases. Nonetheless, while we fixed the threshold at $0.5$ here, we leave open the possibility that tuning it as a hyper-parameter could yield performance improvements.

\begin{table}[htbp]
\caption{Comparison between $\difflift$ (random) and its deterministic variant (thresholding at 0.5).\looseness=-1}
\centering
\small
\resizebox{\textwidth}{!}{\begin{tabular}{@{}lllrrrrr@{}}
\toprule
Domain & TNN & $\partial$lift & NCI1 & NCI109 & ZINC (MAE) & PROTEINS & MUTAG \\
\midrule
Cellular & CWN & deterministic  & $80.62\pm0.75$ & $77.64\pm0.30$ & $1.33\pm5.33$ & $70.54\pm0.00$ & $82.46\pm2.48$ \\
Cellular & CWN & random  & $79.81\pm0.40$ & $80.55\pm0.50$ & $0.17\pm0.00$ & $70.54\pm3.34$ & $85.96\pm4.96$ \\
Hypergraph & UniGCNII & deterministic  & $71.53\pm1.92$ & $69.76\pm6.03$ & $0.70\pm0.01$ & $72.62\pm0.42$ & $75.44\pm2.48$ \\
Hypergraph & UniGCNII & random  & $77.45\pm1.88$ & $75.30\pm1.10$ & $0.56\pm0.03$ & $73.51\pm0.84$ & $89.47\pm4.30$ \\
\bottomrule
\end{tabular}}

\label{tab:deterministic_results}
\end{table}

\subsection{Further comparison against DCM}

\textbf{DCM versus $\partial$lift.} DCM~\citep{battiloro2023latent} introduces a learnable lifting approach specifically for cell complexes through a two-step procedure: first, the $\alpha$-Differentiable Graph Module ($\alpha$-DGM) learns the 1-skeleton (edges) using $\alpha$-entmax sampling to generate sparse, non-regular graphs; then, the Polygon Inference Module (PIM) samples polygons from induced cycles of the learned graph. DCM uses $\alpha$-entmax for both edge and polygon sampling, and trains end-to-end using auxiliary reward-based losses (Eqs. 11 and 13 in their paper) that encourage edges/polygons involved in correct predictions. The method is evaluated exclusively on node classification tasks (homophilic and heterophilic datasets) and is limited to 2-dimensional cell complexes.

Some key differences between DCM and $\partial$Lift include:
\begin{enumerate}[left=8pt,topsep=0pt]
    \item \textit{Domain generality}: our framework applies to hypergraphs, simplicial complexes, and combinatorial complexes, not just cell complexes. Extending DCM to hypergraphs is non-trivial because it relies on cycle-based candidate generation, which is inherently tied to the graph structure and does not naturally translate to hyperedge formation;
    \item \textit{Sampling mechanism and training}: we use Bernoulli sampling with straight-through estimators rather than $\alpha$-entmax, and critically, we do not require auxiliary reward-based losses --- our framework is trained purely with the task loss, making it simpler and better suited for end-to-end learning. Additionally, our hypergraph variant is embarrassingly parallel whereas DCM's two-step procedure is sequential;
    \item \textit{Adaptive cell sizes}: we learn distributions over $k_v$ (neighborhood sizes) allowing adaptive hyperedge/cell cardinalities, while DCM's polygon sizes are constrained by the induced cycles in the learned graph;
    \item \textit{Evaluation scope}: we assess both node and graph classification across 12 datasets with multiple TNN architectures, demonstrating broader applicability.
\end{enumerate}

\textbf{Experiments on point clouds.} We compared $\partial$lift against DCM on point-cloud node classification (graphs with no edges) to provide a direct comparison. \autoref{tab:pointcloud_tnn_comparison} shows results across four datasets. $\partial$lift combined with different TNNs achieves competitive or superior performance: UniGCNII+$\partial$lift achieves 84.97\% on Wisconsin (vs. 71.24\% for DCM), and UniGIN+$\partial$lift achieves 83.78\% on Texas (vs. 62.16\% for DCM), demonstrating substantial gains on heterophilic datasets. On the homophilic dataset Cora, DCM achieves 73.07\% compared to 71.47\% for the best $\partial$lift variant, showing comparable performance. As expected, the performance in heterophilic datasets are higher in this setting compared to homophilic when referring \autoref{tab:cellular_node} and \autoref{tab:hypergraph_node}. These results demonstrate that $\partial$lift's framework provides competitive performance while offering greater flexibility across topological domains and TNN architectures.

\begin{table}[h]
\centering
\caption{Point cloud performance comparison of $\partial$lift across TNNs versus DCM.}
\begin{tabular}{l|cccc}
\hline
\textbf{TNN + $\partial$lift} & \textbf{Citeseer} & \textbf{Texas} & \textbf{Wisconsin} & \textbf{Cora} \\
\hline
UniGCNII & $\textbf{74.73} \textcolor{gray}{\scriptstyle{\pm 0.12}}$ & $81.98 \textcolor{gray}{\scriptstyle{\pm 1.27}}$ & $\textbf{84.97} \textcolor{gray}{\scriptstyle{\pm 2.45}}$ & $70.83 \textcolor{gray}{\scriptstyle{\pm 0.37}}$ \\
CWN & $51.80 \textcolor{gray}{\scriptstyle{\pm 1.67}}$ & $71.17 \textcolor{gray}{\scriptstyle{\pm 1.27}}$ & $75.82 \textcolor{gray}{\scriptstyle{\pm 0.92}}$ & $55.90 \textcolor{gray}{\scriptstyle{\pm 2.45}}$ \\
CXN & $62.03 \textcolor{gray}{\scriptstyle{\pm 1.59}}$ & $81.08 \textcolor{gray}{\scriptstyle{\pm 0.00}}$ & $78.43 \textcolor{gray}{\scriptstyle{\pm 0.00}}$ & $57.43 \textcolor{gray}{\scriptstyle{\pm 1.19}}$ \\
UniGIN & $72.73 \textcolor{gray}{\scriptstyle{\pm 0.61}}$ & $\textbf{83.78} \textcolor{gray}{\scriptstyle{\pm 2.21}}$ & $83.66 \textcolor{gray}{\scriptstyle{\pm 0.92}}$ & $71.47 \textcolor{gray}{\scriptstyle{\pm 1.77}}$ \\
\midrule
DCM & $74.40 \textcolor{gray}{\scriptstyle{\pm 0.45}}$ & $62.16 \textcolor{gray}{\scriptstyle{\pm 5.84}}$ & $71.24 \textcolor{gray}{\scriptstyle{\pm 3.33}}$ & $\textbf{73.07} \textcolor{gray}{\scriptstyle{\pm 0.92}}$ \\
\hline
\end{tabular}
\label{tab:pointcloud_tnn_comparison}
\end{table}

\subsection{Comparison against probabilistic rewiring}

\textbf{IPR-MPNN versus $\partial$lift.} IPR-MPNN~\citep{qian2024probabilistic} introduces probabilistic graph rewiring by connecting original graph nodes to a small set of virtual nodes in an end-to-end differentiable manner. The method uses an upstream MPNN to compute priors $\theta$ for assigning each original node to $k$ virtual nodes (from $m$ total virtual nodes, where $m \ll n$), sampling assignment matrices via differentiable $k$-subset sampling. A downstream MPNN then operates on the augmented graph with message passing among: (1) original nodes to virtual nodes, (2) among virtual nodes (forming a complete subgraph), and (3) among original nodes.

There are relevant distinctions between $\partial$lift and IPR-MPNN: (i) \textit{Explicit vs. implicit structure}: IPR-MPNN implicitly routes long-range information through virtual nodes, while $\partial$lift explicitly constructs higher-order cells (hyperedges, simplices, polygons) that directly encode multi-way interactions; (ii) \textit{Domain flexibility}: IPR-MPNN operates within the graph domain augmented with virtual nodes, whereas $\partial$lift learns liftings to diverse topological domains (hypergraphs, simplicial complexes, cell complexes, combinatorial complexes); (iii) \textit{Sampling strategy}: both use differentiable $k$-subset sampling, but IPR-MPNN samples node-to-virtual-node assignments while $\partial$lift samples which higher-order cells to include in the lifted structure; (iv) \textit{Computational approach}: IPR-MPNN requires managing virtual node features and specialized message-passing between hierarchies, while $\partial$lift integrates directly with standard TNN architectures designed for each topological domain.

\textbf{Results on graph classification.} \autoref{tab:ipr_comparison} compares IPR-MPNN to $\partial$lift combined with various TNNs on five molecular datasets. To ensure a fair comparison, we evaluated IPR-MPNN using our experimental setup, with the data splits described in \autoref{ap:datasets} and a hyperparameter grid similar to the one in \autoref{ap:implementation_details}, with embedding dimensions in $\{32, 64, 128\}$ and network depths in $\{2, 3\}$. All other hyperparameters remained fixed, taken from the configurations in the official IPR-MPNN repository. For instance, on ZINC we do not use edge features.

$\partial$lift achieves substantial improvements on multiple benchmarks: CXN+$\partial$lift reaches 82.08\% on NCI1 and 82.57\% on NCI109 compared to IPR-MPNN's 77.44\% and 77.08\%. On ZINC, both CWN+$\partial$lift and CXN+$\partial$lift achieve 0.17 MAE versus IPR-MPNN's 0.39, representing a 56\% error reduction. UniGCNII+$\partial$lift obtains the highest MUTAG accuracy at 89.47\%. While IPR-MPNN demonstrates the advantages of learnable graph augmentation through virtual nodes, our results show that explicitly learning higher-order topological structures through $\partial$lift can provide complementary benefits, particularly when combined with TNNs designed to exploit these structures.

\color{black}

\begin{table}[htbp!] \centering \caption{Graph classification performance comparison between IPR-MPNN and $\partial$lift combined with various TNNs} \begin{tabular}{lccccc} \toprule \textbf{Method} & \textbf{MUTAG}$\uparrow$ & \textbf{NCI1}$\uparrow$ & \textbf{NCI109}$\uparrow$ & \textbf{PROTEINS}$\uparrow$ & \textbf{ZINC}$\downarrow$ \\ \midrule IPR-MPNN & $70.18\textcolor{gray}{\scriptstyle{\pm 2.48}}$ & $77.44\textcolor{gray}{\scriptstyle{\pm 1.13}}$ & $77.08\textcolor{gray}{\scriptstyle{\pm 0.30}}$ & $73.73\textcolor{gray}{\scriptstyle{\pm 1.93}}$ & $0.39\textcolor{gray}{\scriptstyle{\pm 0.05}}$ \\ \midrule CWN + Cycle & $66.67\textcolor{gray}{\scriptstyle{\pm 12.41}}$ & $76.93\textcolor{gray}{\scriptstyle{\pm 1.18}}$ & $76.71\textcolor{gray}{\scriptstyle{\pm 1.34}}$ & $69.05\textcolor{gray}{\scriptstyle{\pm 2.95}}$ & $0.46\textcolor{gray}{\scriptstyle{\pm 0.01}}$ \\ CWN + $\partial$lift & $85.96\textcolor{gray}{\scriptstyle{\pm 4.96}}$ & $79.81\textcolor{gray}{\scriptstyle{\pm 0.40}}$ & $80.55\textcolor{gray}{\scriptstyle{\pm 0.50}}$ & $70.54\textcolor{gray}{\scriptstyle{\pm 3.34}}$ & $\textbf{0.17}\textcolor{gray}{\scriptstyle{\pm 0.00}}$ \\ \midrule CXN + Cycle & $61.40\textcolor{gray}{\scriptstyle{\pm 2.48}}$ & $72.02\textcolor{gray}{\scriptstyle{\pm 1.69}}$ & $75.01\textcolor{gray}{\scriptstyle{\pm 0.62}}$ & $70.83\textcolor{gray}{\scriptstyle{\pm 1.52}}$ & $0.79\textcolor{gray}{\scriptstyle{\pm 0.02}}$ \\ CXN + $\partial$lift & $84.21\textcolor{gray}{\scriptstyle{\pm 4.30}}$ & $\textbf{82.08}\textcolor{gray}{\scriptstyle{\pm 1.50}}$ & $\textbf{82.57}\textcolor{gray}{\scriptstyle{\pm 0.40}}$ & $69.94\textcolor{gray}{\scriptstyle{\pm 2.10}}$ & $\textbf{0.17}\textcolor{gray}{\scriptstyle{\pm 0.01}}$ \\ \midrule UniGCNII + $k$-hop & $61.40\textcolor{gray}{\scriptstyle{\pm 2.48}}$ & $72.70\textcolor{gray}{\scriptstyle{\pm 0.52}}$ & $72.01\textcolor{gray}{\scriptstyle{\pm 1.55}}$ & $72.92\textcolor{gray}{\scriptstyle{\pm 1.11}}$ & $0.66\textcolor{gray}{\scriptstyle{\pm 0.02}}$ \\ UniGCNII + $\partial$lift & $\textbf{89.47}\textcolor{gray}{\scriptstyle{\pm 4.30}}$ & $77.45\textcolor{gray}{\scriptstyle{\pm 1.88}}$ & $75.30\textcolor{gray}{\scriptstyle{\pm 1.10}}$ & $73.51\textcolor{gray}{\scriptstyle{\pm 0.84}}$ & $0.56\textcolor{gray}{\scriptstyle{\pm 0.03}}$ \\ \midrule UniGIN + $k$-hop & $64.91\textcolor{gray}{\scriptstyle{\pm 2.48}}$ & $65.50\textcolor{gray}{\scriptstyle{\pm 1.99}}$ & $66.97\textcolor{gray}{\scriptstyle{\pm 7.25}}$ & $71.43\textcolor{gray}{\scriptstyle{\pm 0.73}}$ & $1.15\textcolor{gray}{\scriptstyle{\pm 0.01}}$ \\ UniGIN + $\partial$lift & $66.67\textcolor{gray}{\scriptstyle{\pm 6.56}}$ & $64.88\textcolor{gray}{\scriptstyle{\pm 1.09}}$ & $79.74\textcolor{gray}{\scriptstyle{\pm 0.23}}$ & $73.81\textcolor{gray}{\scriptstyle{\pm 1.52}}$ & $0.92\textcolor{gray}{\scriptstyle{\pm 0.05}}$ \\ \bottomrule \end{tabular} \label{tab:ipr_comparison} \end{table}

{




}


\end{document}